%% file: imperceptible_adv_examples.tex
\documentclass[10pt,twocolumn,letterpaper]{article}

\usepackage[affil-it]{authblk}

\usepackage[pagenumbers]{cvpr2025/cvpr} 


\definecolor{cvprblue}{rgb}{0.21,0.49,0.74}
\usepackage[pagebackref,breaklinks,colorlinks,allcolors=cvprblue]{hyperref}

\input{preamble.tex}

\title{Imperceptible Adversarial Examples in the Physical World}

\author[1]{Weilin Xu}
\author[1]{Sebastian Szyller}
\author[1]{Cory Cornelius}
\author[1]{Luis Murillo Rojas}
\author[1]{Marius Arvinte}
\author[2]{Alvaro Velasquez}
\author[3]{Jason Martin\thanks{Contributed to the work while at Intel.} }
\author[1]{Nageen Himayat}

\affil[1]{Intel}
\affil[2]{University of Colorado Boulder}
\affil[3]{HiddenLayer}

\begin{document}
\maketitle

\input{0.abstract.tex}
\input{1.intro.tex}
\input{2.background.tex}
\input{3.global_perturbation_printouts.tex}
\input{4.patch_perturbation_carla.tex}
\input{5.conclusion.tex}
\input{6.ack.tex}

{
    \small
    \bibliographystyle{cvpr2025/ieeenat_fullname}
    \bibliography{imperceptible_adv_examples}
}

\appendix
\counterwithin{figure}{section}
\clearpage
\section{Appendix}
\input{7.appendix}

\end{document}

%% file: preamble.tex
\usepackage{amsmath}
\usepackage{array} 
\usepackage{acronym} 
\usepackage{nth} 
\usepackage{wrapfig} 
\usepackage{tabularx} 
\usepackage{multirow}

\newcolumntype{L}{@{\extracolsep{\fill}}l}
\newcolumntype{R}{@{\extracolsep{\fill}}r}
\newcolumntype{C}{@{\extracolsep{\fill}}c}

\DeclareMathOperator*{\argmin}{argmin}

\acrodef{STE}[STE]{\emph{Straight-through Estimator}}
\acrodef{FGSM}[FGSM]{\emph{Fast Gradient Sign Method}}
\acrodef{PGD}[PGD]{Projected Gradient Descent}
\acrodef{BPDA}[BPDA]{\emph{Backward Pass Differentiable Approximation}}
\acrodef{DNN}[DNN]{\emph{Deep Neural Network}}
\acrodef{DNNs}[DNNs]{\emph{Deep Neural Networks}}
\acrodef{MART}[MART]{\emph{Modular Adversarial Robustness Toolkit}}
\acrodef{EoT}[EoT]{\emph{Expectation over Transformation}}
\acrodef{mAP}[mAP]{\emph{mean Average Precision}}
\acrodef{AP50}[AP50]{\emph{Average Precision 50}}
\acrodef{UE4}[UE4]{\emph{Unreal Engine 4}}

\usepackage{pifont}
\newcommand{\cmark}{\ding{51}}%
\newcommand{\xmark}{\ding{55}}%


%% file: 0.abstract.tex
\begin{abstract}
Adversarial examples in the digital domain against deep learning-based computer vision models allow for perturbations that are imperceptible to human eyes.
However, producing similar adversarial examples in the physical world has been difficult due to the non-differentiable image distortion functions in visual sensing systems.
The existing algorithms for generating physically realizable adversarial examples often loosen their definition of adversarial examples by allowing unbounded perturbations, resulting in obvious or even strange visual patterns.
In this work, we make adversarial examples imperceptible in the physical world using a straight-through estimator (STE, a.k.a. BPDA).
We employ STE to overcome the non-differentiability -- applying exact, non-differentiable distortions in the forward pass of the backpropagation step, and using the identity function in the backward pass.
Our differentiable rendering extension to STE also enables imperceptible adversarial patches in the physical world.
Using printout photos, and experiments in the CARLA simulator, we show that STE enables fast generation of $\ell_\infty$ bounded adversarial examples despite the non-differentiable distortions.
To the best of our knowledge, this is the first work demonstrating imperceptible adversarial examples bounded by small $\ell_\infty$ norms in the physical world that force zero classification accuracy in the global perturbation threat model and cause near-zero ($4.22\%$) AP50 in object detection in the patch perturbation threat model.
We urge the community to re-evaluate the threat of adversarial examples in the physical world.
\end{abstract}

%% file: 1.intro.tex
\section{Introduction}

\ac{DNNs} have been the \textit{de facto} solutions to many computer vision problems thanks to their exceptional accuracy compared with previous methods.
However, \ac{DNNs} are susceptible to \emph{adversarial examples} that degrade performance down to near-zero on most tasks.
It is easy to generate adversarial examples against any \ac{DNNs} using attacks based on gradient descent and backpropagation -- such as \ac{FGSM}~\cite{szegedy2013intriguing} or \ac{PGD}~\cite{madry2017towards}.
On the contrary, attacking visual sensing systems in the physical world is more difficult due to the non-differentiable distortion functions in the imaging pipeline.

Prior work on physically realizable adversarial examples loosen the definition of
adversarial examples by allowing unbounded perturbations confined to a portion of the image~\cite{athalye2018synthesizing,chen2019shapeshifter,jan2019connecting}.
This increases the chance of the perturbations surviving the non-differentiable distortions in the target system.
However, as a side-effect it can lead to obvious or even strange textures in the generated perturbations, making them easier to spot~\cite{chen2019shapeshifter, braunegg2020apricot, wu2020making}.
As a result, adversarial examples do not deter practitioners from deploying \ac{DNNs} in security-sensitive scenarios, such as video surveillance systems~\cite{rezaee2024survey} and autonomous vehicles~\cite{deng2021deep}.
Black Hat Briefings has been reluctant to accept submissions on adversarial examples against visual sensing systems in its AI track, as ``industry hasn't cared much because it doesn't impact most of them''~\cite{bh_ai_track_observations}.

In this work, we challenge the belief that adversarial examples are not a threat thanks to the image distortions in a visual sensing system serving as an effective defense.
Our work focuses on \emph{imperceptible} perturbations in the physical domain that are aligned with the original definition~\cite{szegedy2013intriguing}.
We demonstrate that physically realizable adversarial examples are achievable and a real threat that requires serious consideration.
Hence, practitioners should re-evaluate the security implications of \ac{DNN}-based visual sensing systems and assume that adversarial examples in the physical world can be as effective as in the digital domain.

\begin{figure*}[tp!]
    \centering
    \includegraphics[width=\textwidth]{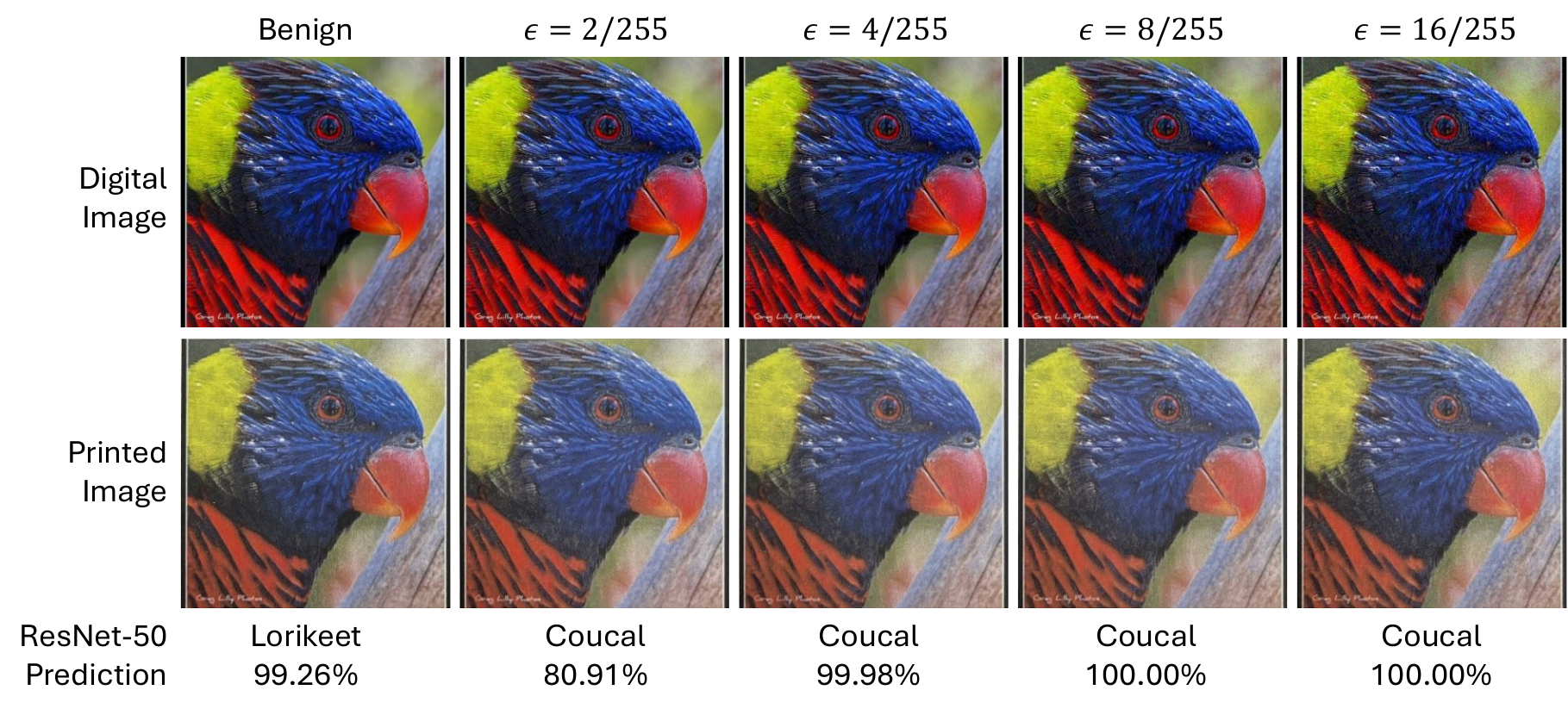}
    \caption{Straight-through estimator (STE, a.k.a. BPDA) combined with \ac{PGD} reliably produces imperceptible adversarial examples in the physical world that fool the target model -- a pre-trained ResNet50 image classifier. We consider global adversarial perturbations bounded by $\ell_\infty=2|4|8|16$. See experimental details in~\Cref{sec:global_perturbation_printouts}.}%
    \label{fig:imperceptible_adversarial_examples_bird}
\end{figure*}

In this work, we develop a novel approach based on \ac{STE} to generate adversarial examples against \ac{DNN}-based visual sensing systems that distort images with non-differentiable functions, such as inaccurate colors in printers, clipping of electrical signals caused by photodiodes or power amplifiers~\cite{buckler2017reconfiguring}.
Our key insight is that we overcome such distortions using non-differentiable components in the forward pass without any approximation, and using the identity function or differentiable rendering in the backward pass to enable backpropagation.
In experiments with the paper printouts and the CARLA simulator, we show that our attack in the physical world has comparable effectiveness to classic attacks in the digital domain.
Additionally, our method can generate $\ell_\infty$ bounded adversarial examples with imperceptible perturbations that are effective in the physical world (as shown in~\Cref{fig:imperceptible_adversarial_examples_bird}).

\subsection{Contributions}
We claim the following contributions:
\begin{enumerate}
    \item We propose a novel attack that uses \ac{STE} and differentiable rendering to generate adversarial examples against \ac{DNN}-based visual sensing systems that contain non-differentiable distortion functions in their imaging pipelines (\Cref{sec:ste_method_global} and~\Cref{sec:ste_method_patch}).
    \item We demonstrate how to use \ac{STE} in the global perturbation threat model to produce imperceptible adversarial examples bounded by $\ell_\infty=4/255$ on paper printouts that force \emph{zero} accuracy on the target model (\Cref{sec:global_perturbation_printouts}).
    \item We combine \ac{STE} with differentiable rendering to produce imperceptible adversarial patches in the CARLA urban driving simulator~\cite{dosovitskiy2017carla} which is built upon \ac{UE4}~\cite{ue} with a non-differentiable 3D renderer. We show that our method produces unbounded adversarial patches that lower AP50 of object detection from $43.29\%$ to $0\%$ and imperceptible adversarial patches bounded by $\ell_\infty=\frac{8}{255}$ that decrease AP50 from $43.29\%$ to $4.22\%$ (\Cref{sec:patch_perturbations_carla}).
\end{enumerate}

%% file: 2.background.tex
\section{Background \& Related Work}

\textbf{Adversarial examples.} Despite substantial research, present-day neural networks remain susceptible to adversarial examples~\cite{athalye2018obfuscated,carlini2019evaluating,carlini2024aligned}, with no computationally efficient defense being widely adopted. Defenses that have empirical success \cite{madry2017towards},
or theoretical performance guarantees \cite{cohen2019certified} typically increase the cost of training or inference by at least one order of magnitude.
An adversary crafts an adversarial example by adding an imperceptible, crafted perturbation to a test sample, and uses it to conduct a \emph{model evasion} attack that fools the model into making a wrong prediction.
This is formalized as solving the following constrained optimization problem:
\begin{equation}
   \argmin_{\delta} \ \mathcal{L}\left(f(x+\delta), y_\textrm{target}\right), \quad \textrm{s.t.} \ \delta \in \mathcal{S},
\label{eq:adv_optimization}
\end{equation}
\noindent where $f$ is the end-to-end processing that produces the estimated output obtained from perturbing the input $x$ with $\delta$, $y_\textrm{target}$ is the desired output, $\mathcal{L}$ is the loss function minimized by the adversary, and $\mathcal{S}$ is a constraint set placed on the perturbation $\delta$.

Generating $\ell_p$ bounded adversarial examples with an imperceptible perturbation in the digital domain using gradient-based approaches is trivial.
However, similar adversarial examples are less effective in the physical space -- if they are printed out, or a camera is used to capture them and feed the frames to the target model~\cite{kurakin2018adversarial}.
This is due to the fact that visual sensing systems distort images by passing them through complex processing pipelines~\cite{buckler2017reconfiguring,mosleh2020hardware} that many attack algorithms do not take into account.

A straightforward way to overcome this is to accurately model the distortion function of the printer (or the screen) and the camera using a differentiable function.
However, this requires non-trivial effort and is usually infeasible, given the complicated modern image capture pipelines~\cite{baqai2005digital}.

Approaches that attempt to overcome the non-differentiability in generating adversarial examples in the physical world can be divided into three groups:
perturbation constraints, \ac{EoT}, and neural rendering.
Recent work on physically realizable adversarial examples often combines techniques from multiple categories~\cite{suryanto2022dta,zhou2024rauca}.

\noindent\textbf{Perturbation constraints.}
Printers and screens do not produce colors with 100\% accuracy~\cite{baqai2005digital}.
Furthermore, imaging sensors do not capture pixels perfectly.
To account for the limitations of rendering and imaging pipelines, and produce effective adversarial examples, we need to lower the discrepancy between a digital image and its corresponding (captured) camera frame.
Early work on physically realizable adversarial examples~\cite{sharif2016accessorize} proposed several physical constraints on adversarial examples:
allowing limited printable colors,
encouraging smooth textures with the total variance term,
and using manual color management to calibrate the printing colors.
Such physical constraints enable adversarial eyeglasses frames that fool a face recognition system.

\noindent\textbf{\ac{EoT}.}
Although modeling the exact distortion functions in the non-differentiable components is non-trivial, we can make adversarial examples robust to distortions in the digital domain.
Hence, the robustness to simulated distortions may generalize well to the actual distortions in the target visual sensing system.
Athalye et al. proposed \ac{EoT} to approximate unknown distortion functions with random image transformations, including rescaling, rotation, lightening, darkening, Gaussian noising, and translation etc~\cite{athalye2018synthesizing}.
\ac{EoT} was later adapted to attacking object detection models in the ShapeShifter work~\cite{chen2019shapeshifter}.

\noindent\textbf{Neural rendering.}
Deep neural networks can be used to approximate the non-differentiable distortion functions.
Jan et al. proposed an image-to-image model that simulates the distortions of an imaging pipeline that consists of a printer and a camera~\cite{jan2019connecting}.

Nevertheless, all these approaches increase the adversarial perturbation budget.
Hence, it becomes difficult to generate $\ell_p$ bounded adversarial examples with imperceptible perturbations in the physical world.
Many attack algorithms loosen the conventional $\ell_p$ constraint on pixel values in favor of unbounded perturbations.
As a result, existing adversarial examples that are effective in the physical world often contain obvious or even strange visual patterns,
which leads to many defense solutions dedicated to detecting such abnormal textures~\cite{liang2021we,jutras2022detecting}.

%% file: 3.global_perturbation_printouts.tex
\section{Imperceptible global perturbations}
\label{sec:global_perturbation_printouts}

In this section, we first explain how to use \ac{STE} to overcome the obstacle of non-differentiable distortions in the imaging pipeline.
Then, we empirically show (using paper printouts) that \ac{STE} is effective in producing imperceptible adversarial examples in the physical world under the global perturbation threat model.

\subsection{\ac{STE} for non-differentiable distortions}
\label{sec:ste_method_global}

\begin{figure*}[tp]
    \centering
    \includegraphics[width=0.8\textwidth]{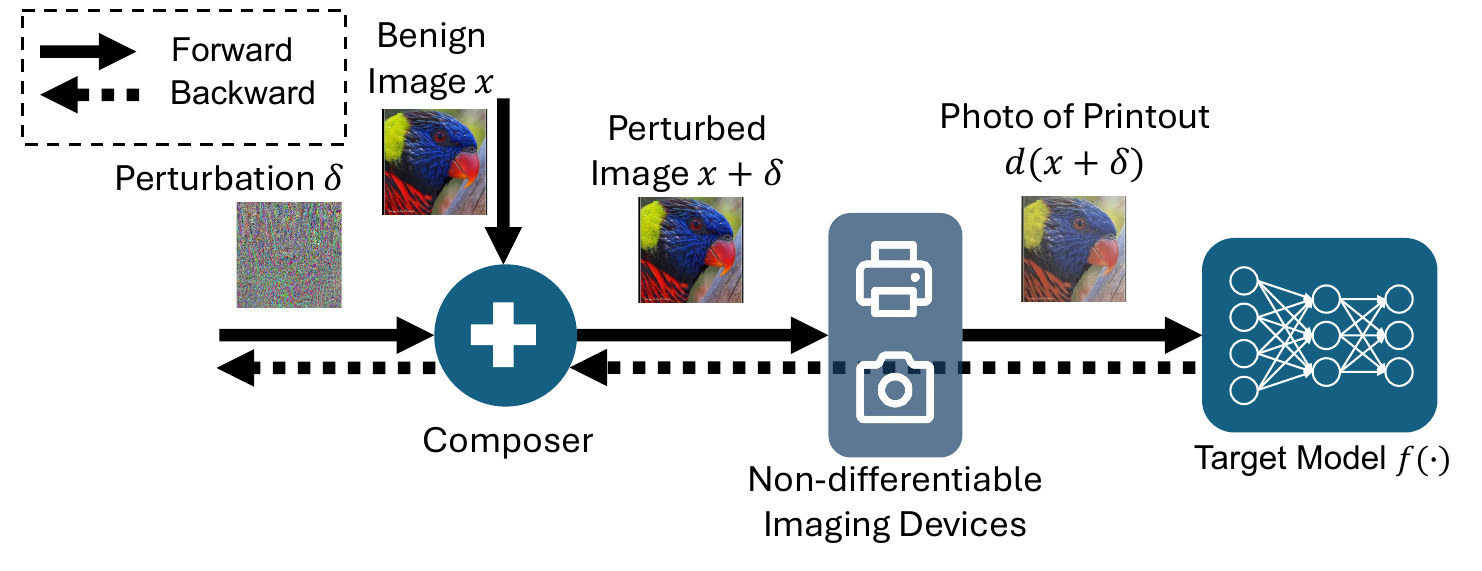}
    \caption{We use \ac{STE} (a.k.a. BPDA) to cross the non-differentiable barrier of the imaging pipeline under the global perturbation threat model. The key is to calculate the accurate loss function value with the non-differentiable distortion functions in the forward pass, but use the identity function in the backward pass to estimate the gradient.}%
    \label{fig:ste_global_perturbation}
\end{figure*}

We formalize a visual sensing system as follows:
given a target model $f(\cdot)$,
an adversary can add an $\ell_\infty$-bounded perturbation $\delta$ to an image $x$ in the digital domain.
The adversary must use some device, such as a printer, to render the digital image $x+\delta$ in the physical world.
The visual sensing system takes a photo of the rendered image, extracts the image and feeds it to the target model.
If we model the distortions on the digital image $x+\delta$ throughout the printer-camera-extractor pipeline as $d(\cdot)$,
the end-to-end visual sensing system can be expressed as:
\begin{equation}
    y = f(d(x+\delta)),
\end{equation}
\noindent where $y$ is the final predicted output. 

The adversary intends to influence $y$ by finding an imperceptible perturbation $\delta$. 
If $d(\cdot)$ is continuous and differentiable, generating adversarial examples against the system should be as easy as an attack in the digital domain.
However, the camera is a non-differentiable function that distorts captured frames due to the imperfections of the sensor and the lens.
The function of the printing device is not differentiable either, and it distorts $x+\delta$ by outputting inaccurate pixel values.
Thus, the adversary can not backpropagate through $d(\cdot)$ to get the gradient of $\delta$ needed to generate adversarial examples.

A straightforward solution is to implement $d(\cdot)$ in a fully differentiable manner.
But it is usually deemed too expensive or even infeasible~\cite{kato2020differentiable}.

Instead, we use \ac{STE}~\cite{bengio2013estimating} to model the non-differentiable distortions, as shown in~\Cref{fig:ste_global_perturbation}.
STE uses the non-differentiable output in the forward pass to obtain the exact loss function value but the identity function in the backward pass to approximate its gradient with respect to the input.
If the non-differentiable distortions lead to a similar image, like what a printer or a camera does, the approximation error should be small.
Intuitively, if we brighten one pixel in the digital image, the printer and the camera would brighten the corresponding pixel in the photo too.
Thus, the sign of the approximate gradient is always accurate, even though the magnitude deviates.
Coincidentally, many $\ell_\infty$ attack algorithms, such as \ac{FGSM} and \ac{PGD}, only use the gradient sign during the optimization.
Therefore, \ac{STE} is the optimal method to overcome non-differentiable distortions for those gradient-sign based attacks.

We use the stop-gradient ($\texttt{sg}[\cdot]$) operation in automatic differentiation systems to construct a differentiable imaging pipeline with distortions as: 
\begin{equation}
\label{eq:ste_approximate}
\begin{aligned}
f(d(x + \delta)) & = f((x+\delta) + \texttt{sg}[d(x+\delta) - (x+\delta)]) \\
                 & = f(x+\delta + C)
\end{aligned}
\end{equation}
\noindent, where $C$ is a constant of the difference between the distorted image and the digital one.

The method is also known as \ac{BPDA} in the adversarial machine learning literature,
and was used to break many defenses relying on non-differentiable filtering functions in the digital domain~\cite{athalye2018obfuscated}.
In the next sections, we show that STE allows us to overcome non-differentiable image distortions when generating adversarial examples in the physical world.

\subsection{Experimental Setup}

\begin{figure}[tp]
    \centering
    \includegraphics[width=0.5\textwidth]{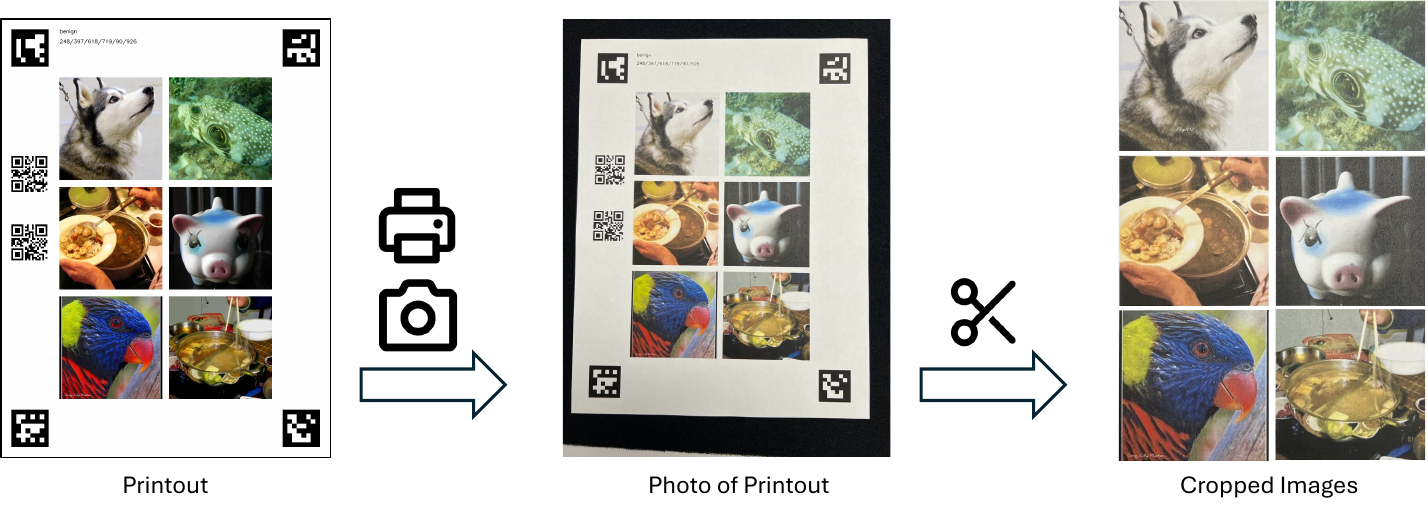}
    \caption{We follow the experiments by Kurakin et al.~\cite{kurakin2018adversarial} to validate the power of \ac{STE} in producing imperceptible adversarial examples in the physical world: 1. generate a digital printout of six square images; 2. print it out and take a photo of the paper; 3. perform perspective transformation and crop out the square images that are being fed to the target model.}%
    \label{fig:exp_setup_printouts}
\end{figure}

We follow the experiment by Kurakin et al.~\cite{kurakin2018adversarial} to validate the power of \ac{STE} in producing imperceptible adversarial examples in the physical world.
We make printouts of six square images with a printer, take photos and crop out the square images to feed the target model,~\Cref{fig:exp_setup_printouts}.

\noindent \textbf{Dataset}: We randomly draw six images of different categories whose edges are not shorter than 300 pixels from the ImageNet ILSVRC validation set~\cite{deng2009imagenet}. We make a center crop of $300 \times 300$ for all images, as shown in \Cref{fig:exp_setup_printouts}.

\noindent \textbf{Target model}: We use the ResNet-50 model from torchvision that is trained on the ImagNet ILSVRC dataset.

\noindent \textbf{Attack algorithms}: We use both the single-step \ac{FGSM} and the 12-step \ac{PGD} with the step size $1/255$.
Both are untargeted attacks trying to increase the cross entropy loss. 
We consider $\ell_\infty$ bounds of 2, 4, 8 and 16 out of 255, which usually produce imperceptible perturbations in the physical world.
We understand that they are by no means the optimal attack algorithms, but they are sufficient to force misclassifications in our experiment.
We compare the performance of all attacks with and without \ac{STE}.

\noindent \textbf{Printout}: As is shown in \Cref{fig:exp_setup_printouts}, we use ArUco markers~\cite{opencv_aruco_markers} and QR codes to automate the extraction of images from photos with perspective transformation and cropping.
We make a total of 113 printouts in the experiment \footnote{113 = 1 benign + 4 epsilons * 2 modes * (1 PGD random init + 12 PGD iters + 1 FGSM iter)}.

\noindent \textbf{Printer}: We use a Canon imageCLASS MF642Cdw color laser printer with resolution 600x600 dpi to make all printouts.
Since it is not a photo-grade printer, we expect to see inaccurate colors and other defects in the printouts.
In fact, the printing distortion is the major obstacle to generating adversarial examples.

\noindent \textbf{Camera}: We use the 12MP Wide Camera in an iPhone 13 Pro to take 113 photos of printouts manually. The handheld operation does not guarantee exactly the same perspective for each photo, which adds more difficulty to the adversary in modeling the distortions.

\subsection{Results}

\begin{table}[]
\centering
\caption{A comparison of FGSM and PGD attacks with and without \ac{STE} in generating imperceptible adversarial examples in the physical world (the printer-camera-extractor pipeline in~\Cref{fig:exp_setup_printouts}).
The goal of the untargeted adversary is to lower the number of correct predictions.
The pure digital FGSM and PGD attacks without \ac{STE} generate perturbations that are destroyed by the printing distortions, resulting in  images being correctly classified in the physical world.
\ac{STE} consistently helps FGSM and PGD generate effective adversarial examples in the physical world, leading to zero accuracy in many cases.
STE-augmented attacks perform poorly in the digital domain because the attacks compute the loss function with the physical world predictions (\Cref{eq:ste_approximate}).}
\label{tab:results_printouts}
\begin{tabularx}{0.4\textwidth}{LCCRR}
\toprule
\multirow[c]{2}{*}{\textbf{Method}} & \multirow[c]{2}{*}{\textbf{$\epsilon$}} & \multirow[c]{2}{*}{\textbf{STE}} & \multicolumn{2}{c}{\textbf{Correct Predictions}$\downarrow$}\\
\cmidrule(lr){4-5}
& & & \textbf{Digital} & \textbf{Physical} \\
\toprule
Benign & 0       & -   & 5/6       & 5/6        \\
\midrule
\multirow[c]{8}{*}{FGSM}   & \multirow[c]{2}{*}{2}       & \xmark   & 1/6       & 4/6        \\
       &         & \cmark   & 4/6       & 4/6        \\
       \cmidrule(lr){2-5}
       & \multirow[c]{2}{*}{4}       & \xmark   & 1/6       & 3/6        \\
       &         & \cmark   & 3/6       & \textbf{2/6}        \\
       \cmidrule(lr){2-5}
       & \multirow[c]{2}{*}{8}       & \xmark   & 0/6       & 3/6        \\
       &         & \cmark   & 3/6       & \textbf{1/6}        \\
       \cmidrule(lr){2-5}
       & \multirow[c]{2}{*}{16}      & \xmark   & 2/6       & 2/6        \\
       &         & \cmark   & 2/6       & \textbf{0/6}        \\
\midrule
\multirow[c]{8}{*}{PGD}    & \multirow[c]{2}{*}{2}       & \xmark   & 0/6       & 3/6        \\
       &         & \cmark   & 5/6       & \textbf{2/6}        \\
       \cmidrule(lr){2-5}
       & \multirow[c]{2}{*}{4}       & \xmark   & 0/6       & 3/6        \\
       &         & \cmark   & 3/6       & \textbf{0/6}        \\
       \cmidrule(lr){2-5}
       & \multirow[c]{2}{*}{8}       & \xmark   & 0/6       & 3/6        \\
       &         & \cmark   & 3/6       & \textbf{0/6}        \\
       \cmidrule(lr){2-5}
       & \multirow[c]{2}{*}{16}      & \xmark   & 0/6       & 2/6        \\
       &         & \cmark   & 3/6       & \textbf{0/6}       \\
\bottomrule
\end{tabularx}
\end{table}

We present the classification performance of the target model on the six images under various attacks in~\Cref{tab:results_printouts}.
Firstly, the image distortions in the physical world do not affect the classification performance on benign images --  the same five images out of six are correctly classified.
However, the same distortions destroy the digital adversarial examples as shown in prior work~\cite{kurakin2018adversarial}.
When we print the digital adversarial examples out on paper, the number of correct predictions increases.
The target model is able to make $2/6$ correct predictions with the strongest digital attack bounded by $\epsilon=16/255$.

In contrast, we learn from the highlighted numbers in the ``Physical'' column of~\Cref{tab:results_printouts} that \ac{STE} strengthens all attacks in the physical world except for FGSM bounded by $\epsilon=2/255$.
In particular, we are able to force zero accuracy when we combine \ac{STE} with the single-step FGSM attack bounded by $\epsilon=16/255$.
In the case of the iterative PGD attacks, zero accuracy is achieved with $\epsilon$ as small as $4$.
We are not much concerned with the fact that we cannot achieve zero accuracy with the PGD attack bounded by $\epsilon=2/255$ for two reasons.
Firstly, PGD attacks bounded by larger $\epsilon$ result in zero accuracy with adversarial perturbations that are also imperceptible to humans, as shown in~\Cref{fig:imperceptible_adversarial_examples_bird} (and more in~\Cref{fig:pgd_ste_printed_all}).
Secondly, we expect to get higher success rates if we combine \ac{STE} with stronger attack algorithms (e.g. Auto-PGD~\cite{croce2020reliable}) or better loss functions (e.g. the hinge loss in CW attacks~\cite{carlini2017towards}).

We design the loss function to optimize attacks in the physical world, and hence, the \ac{STE}-augmented attacks perform worse in the digital domain.

\begin{figure*}[htp]
    \centering
    \begin{subfigure}[b]{0.24\textwidth}
        \centering
        \includegraphics[width=\textwidth]{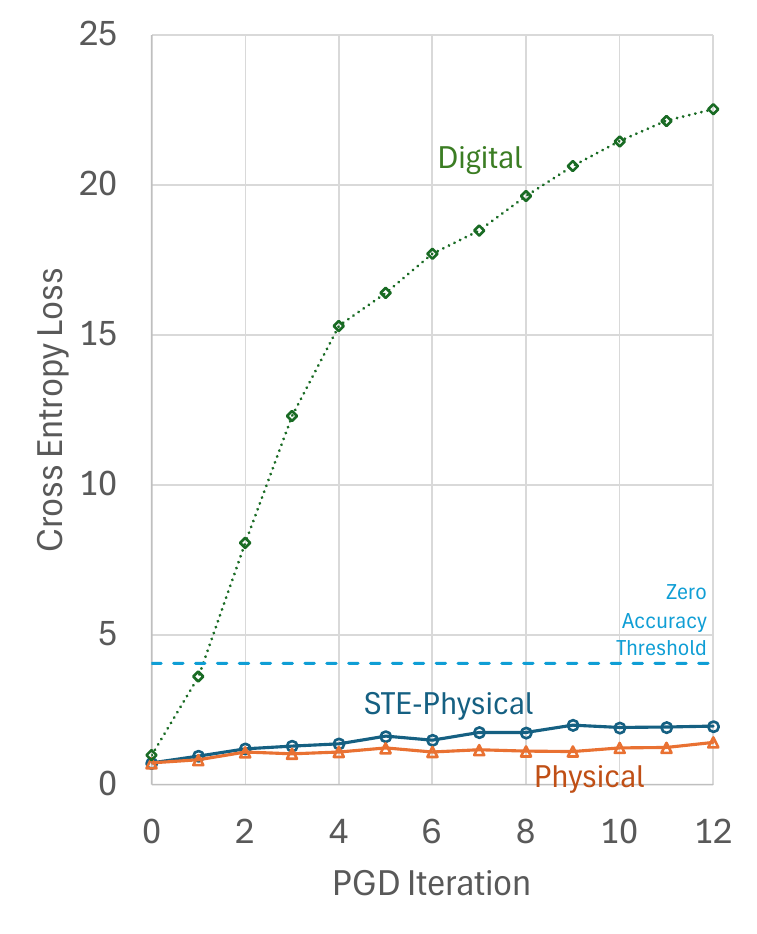}
        \caption{$\epsilon = 2 / 255 $}%
        \label{fig:pgd_loss_eps2}
    \end{subfigure}\hfill%
    \begin{subfigure}[b]{0.24\textwidth}
        \centering
        \includegraphics[width=\textwidth]{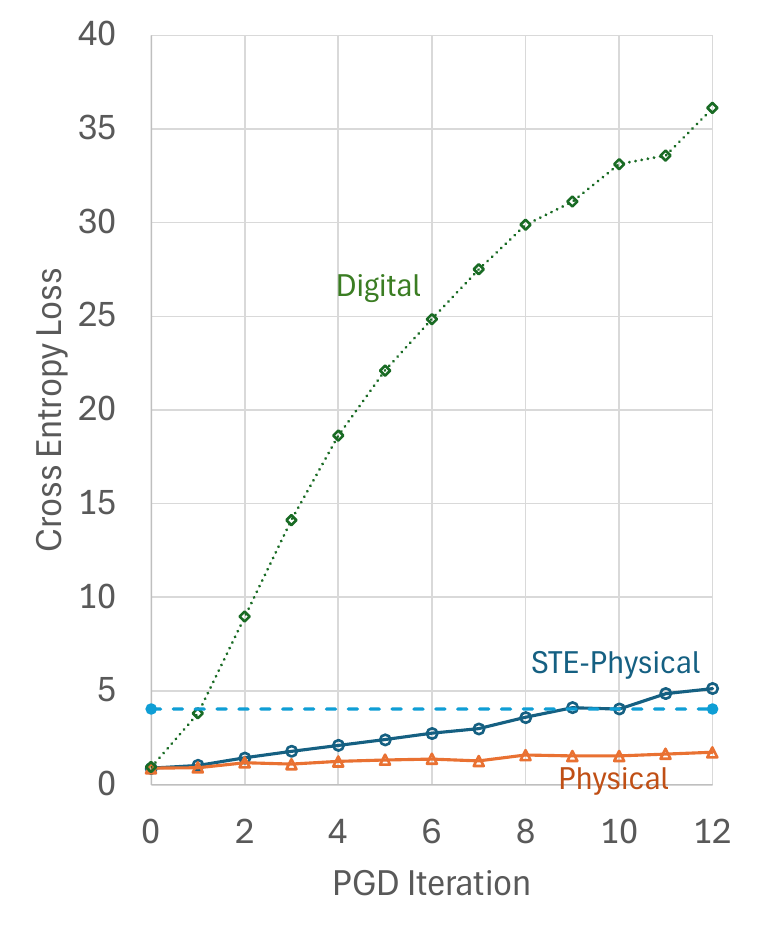}
        \caption{$\epsilon = 4 / 255 $}%
        \label{fig:pgd_loss_eps4}
    \end{subfigure}\hfill%
    \begin{subfigure}[b]{0.24\textwidth}
        \centering
        \includegraphics[width=\textwidth]{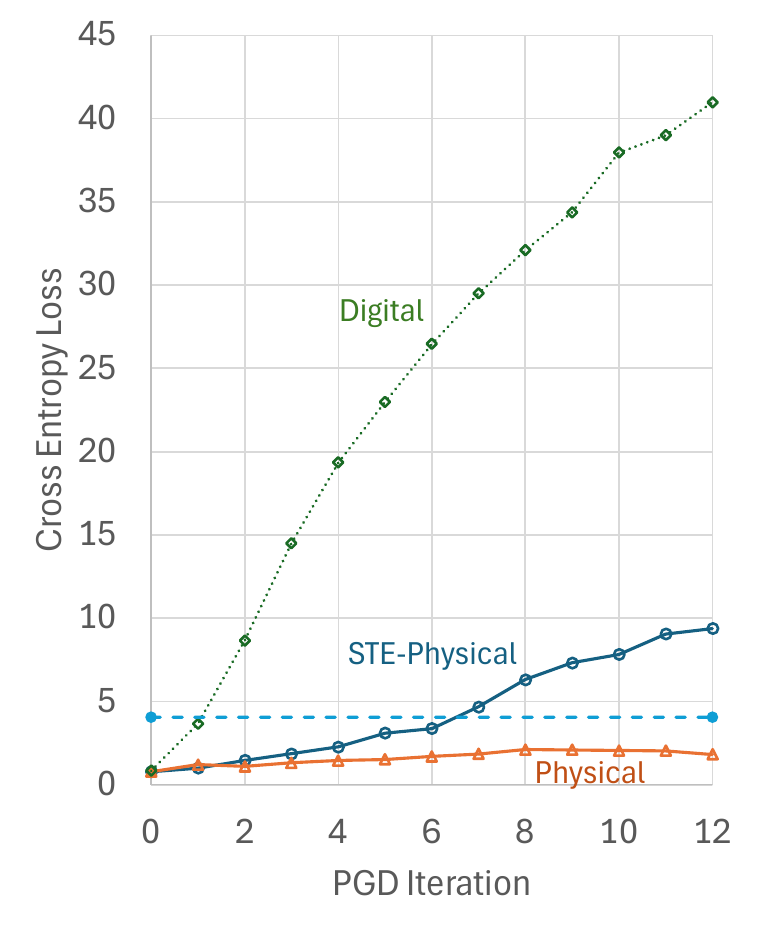}
        \caption{$\epsilon = 8 / 255 $}%
        \label{fig:pgd_loss_eps8}
    \end{subfigure}\hfill%
    \begin{subfigure}[b]{0.24\textwidth}
        \centering
        \includegraphics[width=\textwidth]{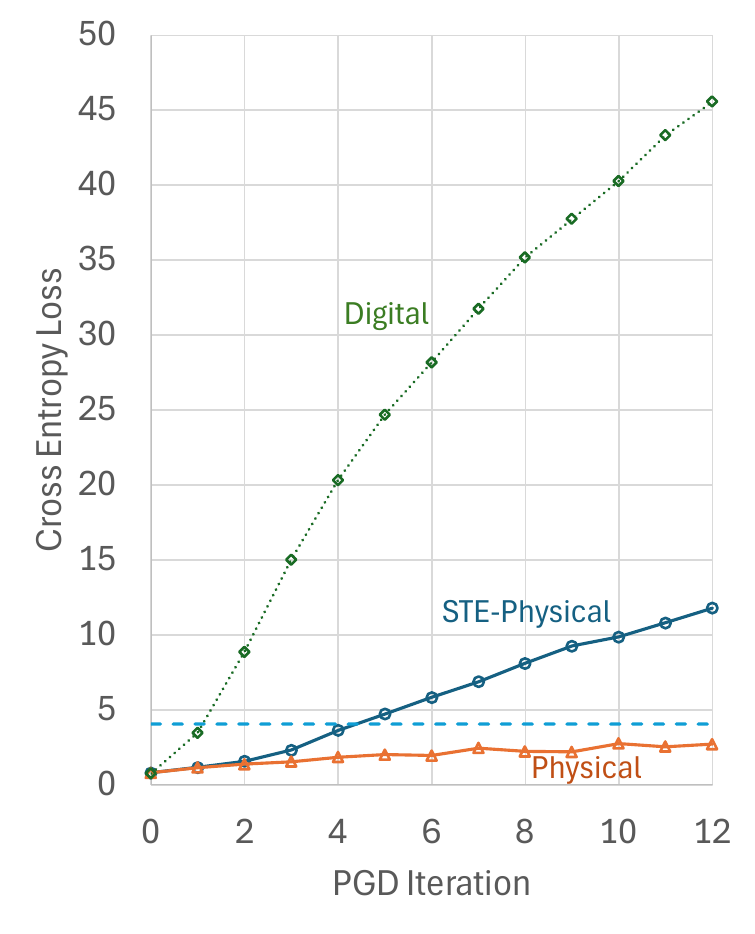}
        \caption{$\epsilon = 16 / 255 $}%
        \label{fig:pgd_loss_eps16}
    \end{subfigure}
    \caption
    {The loss curves of untargeted PGD attacks bounded by four $\ell_\infty$ norms respectively.
    PGD in the digital domain (green dotted curve) is always effective in finding adversarial perturbations that increase the cross entropy loss with respect to the ground truth.
    However, the perturbations are not as effective if we print them out in the physical world (orange curve with triangle markers).
    \ac{STE} helps the PGD optimization of adversarial perturbations in the non-differentiable physical environment (blue curve with circle markers). 
    The STE-augmented PGD attack manages to cross the empirical threshold of loss (cyan dashed line) to get $0\%$ accuracy in the physical world with imperceptible adversarial perturbations bounded by $\ell_\infty$ norms as small as $=4/255$, as is shown in \Cref{fig:imperceptible_adversarial_examples_bird} and \Cref{tab:results_printouts}.}%
    \label{fig:printouts_pgd_loss_curves}
\end{figure*}

We also examine the loss curves to understand how \ac{STE} improves optimization of perturbations iteratively in the physical world (c.f.~\Cref{fig:printouts_pgd_loss_curves}).
The iteration zero on the X-axis corresponds to the random initialization of perturbations before the twelve optimization steps.
The cyan dashed horizontal lines represent the empirical minimum value of cross entropy that links to zero accuracy on the six images.
By reading the green dotted curves, we learn that the PGD attacks in the digital domain take no more than two iterations to achieve zero accuracy.
However, the orange curves with triangle markers below the dashed threshold confirm that the further improved digital perturbations do not transfer to the physical world, as shown earlier in~\Cref{tab:results_printouts}.
The blue curves with circle markers in the middle ground represent our \ac{STE}-augmented PGD attacks.
We find that it only takes five iterations to produce imperceptible adversarial perturbations bounded by $\epsilon=16/255$ that degrade the accuracy of the target model to zero.
Increasing the step size of $1/255$ can further speed up finding effective perturbations, since FGSM succeeds with one step of size $16/255$.

\subsection{Discussion}
Even though we are able to produce imperceptible adversarial examples that result in zero accuracy, we recognize that there is a large gap between the power of the digital attack and our physical attack, as shown in the loss curves in~\Cref{fig:printouts_pgd_loss_curves}.
We identify two possible reasons behind the phenomenon.
Firstly, our printout experiments do not always perfectly crop out the distorted images from photos (see examples in~\Cref{fig:pgd_ste_printed_all}).
Any pixel misalignment would break the assumption of \ac{STE}, resulting in non-optimal solutions to gradient-sign attacks.
Secondly, the image distortions in the physical world constrain the pixel value space, which limits the capability of adversaries in the physical world.
Although existing works, including this paper, fail to find a distortion that can defend computer vision models against adversarial examples in the digital domain,
researchers may get inspired by the physical distortions in search of adversarial example solutions.

%% file: 4.patch_perturbation_carla.tex
\section{Imperceptible patch perturbations}
\label{sec:patch_perturbations_carla}

While global perturbation is the most studied threat model in adversarial machine learning,
patch perturbations are more realistic in the physical world.
In this section, we extend the \ac{STE} method to work with the patch perturbation threat model,
and demonstrate it in the CARLA simulator.

\subsection{\ac{STE} extension with differentiable rendering}
\label{sec:ste_method_patch}

\begin{figure*}[tp]
    \centering
    \includegraphics[width=0.85\textwidth]{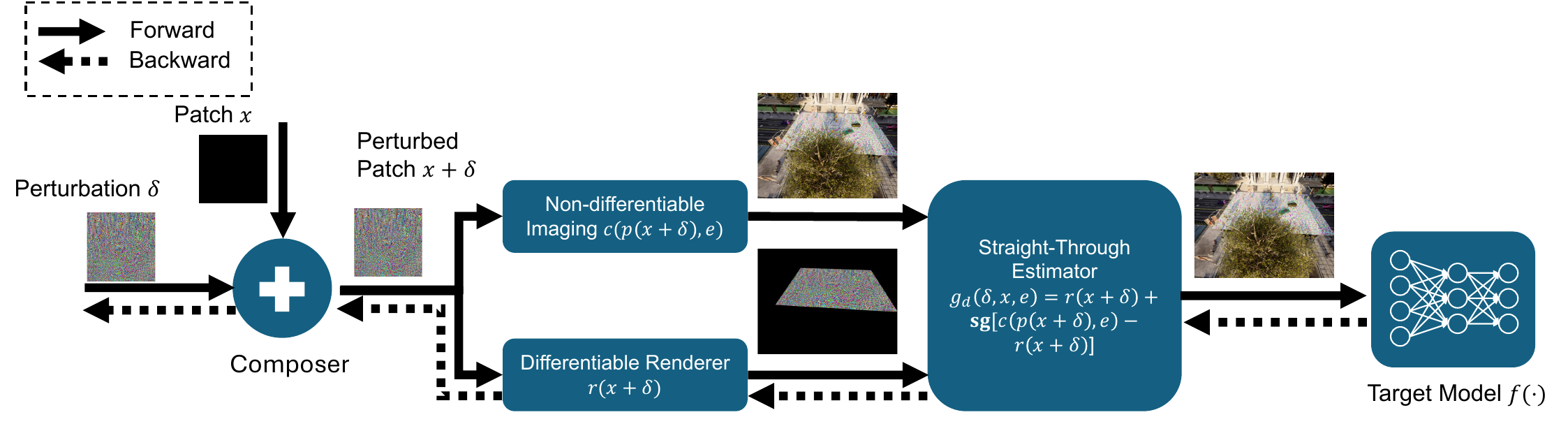}
    \caption{\ac{STE} combined with differentiable rendering overcomes non-differentiability in the patch threat model. }%
    \label{fig:ste_method_patch}
\end{figure*}

We formalize the visual sensing system under the adversarial patch threat model:
given a target model $f(\cdot)$ and a camera $c(\cdot)$,
an adversary can add an $\ell_\infty$-bounded perturbation $\delta$ to a patch image $x$ in the digital domain.
The adversary must use a printing device $p(\cdot)$ to render the perturbed patch in the physical world,
and it cannot change the surrounding environment $e$.
The end-to-end visual sensing system can be expressed as:
\begin{equation}
    y = f(c(p(x+\delta), e)),
\end{equation}

To overcome non-differentiability, we replace the non-differentiable $c(p(x+\delta), e)$ with $c_d(\delta, x, e)$ such that $c_d(\delta, x, e)$ is differentiable with respect to $\delta$ and use gradient-based methods to optimize the perturbation.
As is shown in \Cref{fig:ste_method_patch}, we use a differentiable rendering engine $r(\cdot)$ and a stop-gradient ($\texttt{sg}[\cdot]$) operation to construct a differentiable imaging pipeline as: 
\begin{equation}
\label{eq:diff_rendering_approximate}
\begin{aligned}
g_d(\delta, x, e) &= r(x+\delta) + \texttt{sg}[c(p(x+\delta), e) - r(x+\delta)] \\
                  &= r(x+\delta) + C
\end{aligned}
\end{equation}

It is important to note that: 
\begin{enumerate}
    \item $g_d(\delta, x, e)$ outputs the same as $c(p(x+\delta), e)$ does in the forward pass, because $r(x+\delta)$ cancels out $-r(x+\delta)$ numerically.
    \item $g_d(\delta, x, e)$ is differentiable with respect to $\delta$ through differentiable rendering $r(x+\delta)$.
    \item The differentiable renderer $r(\cdot)$ ignores the surrounding environment $e$.
    \item The difference between the non-differentiable $c(p(x+\delta), e)$ and differentiable $r(x+\delta)$ is detached as a constant $C$ for modern autograd engines~\cite{paszke2019pytorch} in the backward pass by the $\texttt{sg}[\cdot]$ operation.
\end{enumerate}

Therefore, we can approximate the derivative of $f(c(p(x+\delta),e))$ at the point $\hat{\delta}$ as:
\begin{equation}
\begin{aligned}
    \nabla_\delta f(c(p(x+\delta),e)) | _{\delta=\hat{\delta}} &\approx \nabla_\delta f(g_d(\delta,x,e))|_{\delta=\hat{\delta}} \\
    &= \nabla_\delta f(r(x+\delta)+C)|_{\delta=\hat{\delta}}. \\
\end{aligned}
\end{equation}

As long as $r(\cdot)$ renders $x+\delta$ at the same location as in $c(p(x+\delta), e)$, we have the property $c(p(x+\delta), e) \approx r(x+\delta)$ for all pixels of the perturbed image $x+\delta$, thus $\nabla_\delta c(p(x+\delta), e) \approx \nabla_\delta r(x+\delta)$.
The adversary does not care whether $g_d(\delta, x, e)$ is differentiable with respect to $e$, as the adversary has no control over the surrounding environment. 

\subsection{Experimental setup}
We follow the setup in the DARPA GARD~\cite{gard} \nth{7} evaluation to evaluate adversarial patches in the CARLA simulator~\cite{armory_obj_detect_patch_attack}.
One big change we introduce in this paper is to render adversarial patches in CARLA for evaluation, 
while the GARD setup only evaluates the digitally composed patches.
The GARD setup allows unbounded adversarial patches, because the patches only take a small portion of the whole scene.
In this paper, we also evaluate $\ell_\infty$ attacks bounded by small $\epsilon$ to produce imperceptible adversarial patches.

\begin{figure*}[tp]
    \centering
    \includegraphics[width=\textwidth]{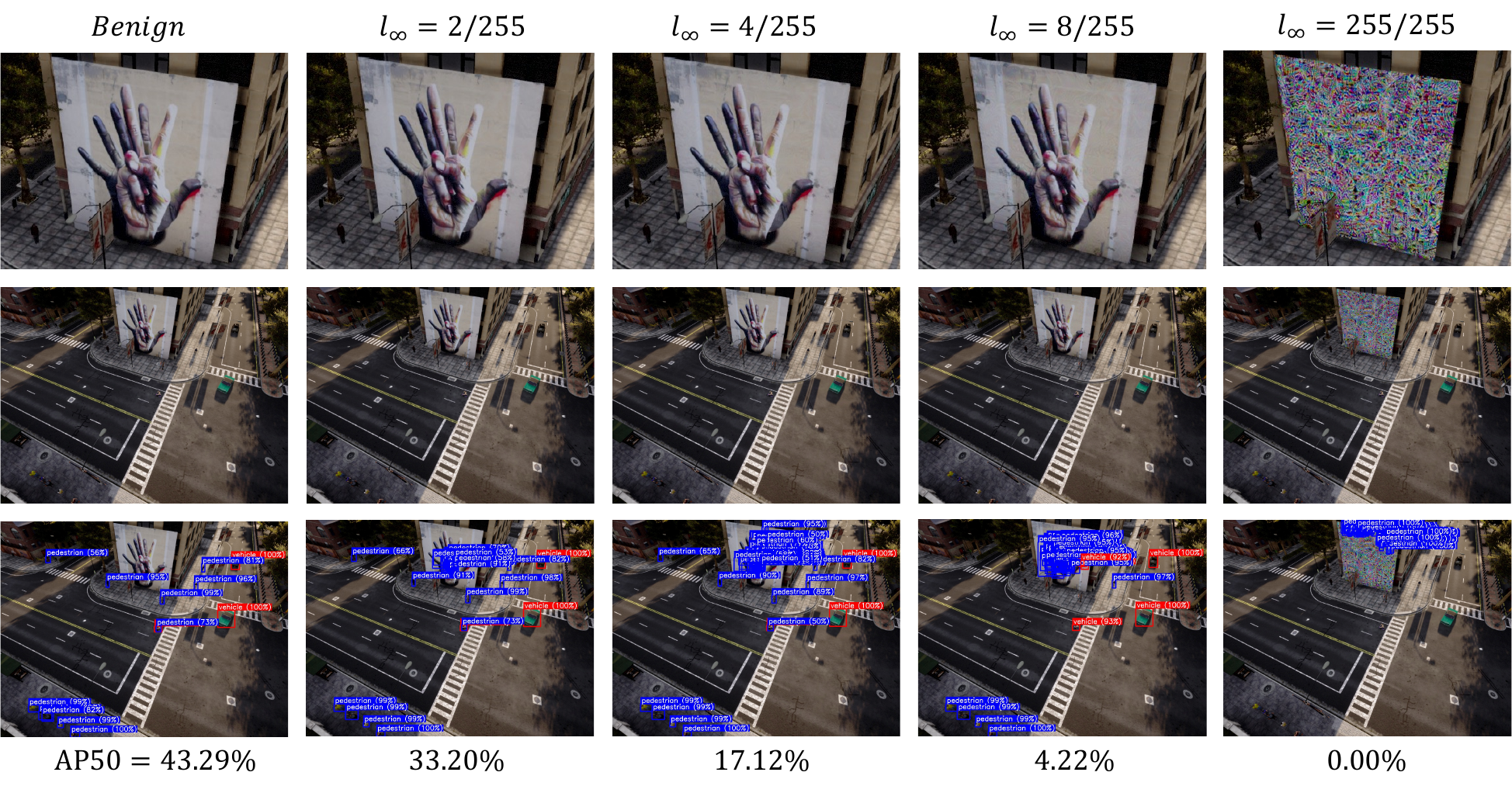}
    \caption{Our method produces imperceptible adversarial examples in CARLA, as the three center columns show. We add the benign images in the left column and the unbounded adversarial examples in the right column for comparison. We also highlight the rectangular patches from the full scenes in the first row, so that it is easier for readers to examine the perturbations.}%
    \label{fig:carla_imperceptible_patches}
\end{figure*}

\noindent \textbf{Dataset}: We use a self-collected dataset that is similar to the DARPA GARD \nth{7} evaluation dataset for object detection, which consists of $20$ images with green rectangular screens that adversaries can perturb to influence the target model (one in \Cref{fig:carla_imperceptible_patches}, more in~\Cref{fig:carla_eval7_dataset}).
Due to the lack of essential CARLA metadata in the released GARD dataset, we have to recreate the scenes of every image using the OSCAR Datagen Toolkit~\cite{cornelius2023oscar} so that we can render custom textures on the green screens in CARLA.
We also use one city art asset (in \Cref{fig:carla_imperceptible_patches}) from CARLA as the starting image to demonstrate imperceptible adversarial perturbations bounded by small $\ell_\infty$ norms.
The rectangular patch can be partially obstructed by other objects (e.g. trees, vehicles).

\noindent \textbf{Target model}: We use the Faster R-CNN object detection model provided in Armory~\cite{Slater_armory_2023} as the target model.
This model is pre-trained on the MS-COCO dataset~\cite{lin2014microsoft} and fine-tuned on the CARLA training dataset offered in Armory to detect only two classes: vehicle and pedestrian.
The model weights file is publicly available online~\cite{gard_eval_armory_carla_model_weights}.

\noindent \textbf{Metrics}: We report the average precision at IoU threshold of $50\%$ (AP50) as the major performance metric for object detection in the experiments.
In addition, we report six TIDE errors to help understand how the model performs in object detection behind the AP50 numbers~\cite{tide-eccv2020}.

\noindent \textbf{Attack}: We implement the iterative attacks using the \ac{MART}~\cite{xu2023modular} framework.
Following the parameters of the example attack in ARMORY~\cite{armory_obj_detect_patch_attack},
we use the Adam optimizer to maximize the training loss for $500$ iterations with the learning rate $\frac{12.75}{255}$.
We use lower learning rates and fewer iterations for the imperceptible adversarial perturbation experiments.
We name the attacks as ``[Optimizer]-[Iterations]-[Learning Rate]'' in ~\Cref{tab:carla_aggregated_results}.

\noindent \textbf{Non-diferentiable renderer}: CARLA uses \ac{UE4} as the renderer which supports complex lighting and materials. 
The nearby objects and the weather in UE4 also affect the rendering result of the adversarial patch.

\noindent \textbf{Differentiable renderer}: Our method does not require a full-function differentiable renderer that is comparable with \ac{UE4}. Since we only render 2D rectangular patches, we use the \emph{perspective transformation} function in PyTorch~\cite{pytorch_perspective_transform}.

\subsection{Results}

In~\Cref{fig:carla_imperceptible_patches} we present one example scene out of 20 in CARLA (and more in~\Cref{fig:carla_ste_eps0_montage,fig:carla_ste_eps0_pred_montage,fig:carla_ste_eps2_montage,fig:carla_ste_eps2_pred_montage,fig:carla_ste_eps4_montage,fig:carla_ste_eps4_pred_montage,fig:carla_ste_eps8_montage,fig:carla_ste_eps8_pred_montage,fig:carla_ste_eps255_montage,fig:carla_ste_eps255_pred_montage,});
we render both benign and adversarial patches for comparison.
The adversarial patches shown in the three center columns cause many hallucinations to the target object detection model but remain imperceptible.

\begin{table*}[]
\centering
\begin{tabularx}{0.8\textwidth}{LLCRRRRRRR}
\toprule
\multirow[c]{2}{*}{\textbf{\shortstack{Adversary \\Budget}}} & \multirow[c]{2}{*}{\textbf{\shortstack[l]{Patch / \\Attack}}}  & \multirow[c]{2}{*}{\textbf{STE}} & \multicolumn{6}{c}{\textbf{TIDE Error Counts}}        &  \multirow[c]{2}{*}{\textbf{AP50}$\downarrow$}     \\
\cmidrule(lr){4-9}
 &          &   & Class             & Box & Other & Dup. & Bg. & Missed &   \\
\toprule
\multirow[c]{3}{*}{\shortstack{0 \\(Benign)}}           & No Patch & -   & 1                 & 13  & 1     & 1         & 10         & 303    & 44.81\% \\
                 & Green Screen    & -   & 2                 & 10  & 1     & 1         & 10         & 302    & 44.74\% \\
                 & City Art        & -   & 1                 & 9   & 1     & 1         & 15         & 306    & 43.29\% \\
\midrule
$\ell_\infty=2$            & Adam-200-1  & \cmark   & 0                 & 16  & 2     & 0         & 44         & 342    & 33.20\%  \\
$\ell_\infty=4$            & Adam-200-2  & \cmark   & 0                 & 20  & 5     & 0         & 86         & 391    & 17.12\% \\
$\ell_\infty=8$            & Adam-200-2  & \cmark   & 0                 & 19  & 28    & 0         & 110        & 436    & \textbf{4.22\%}  \\
\multirow[c]{2}{*}{Unbounded}        & Adam-500-12.75        & \xmark   & 0                 & 19  & 3     & 1         & 16         & 314    & 38.73\% \\
                 & Adam-500-12.75  & \cmark   & 0                 & 30  & 13    & 0         & 157        & 476    & \textbf{0.00\%}    \\
\bottomrule
\end{tabularx}
\caption{STE combined with differentiable rendering produces imperceptible adversarial patches bounded by $\ell_\infty$ norms in CARLA.}
\label{tab:carla_aggregated_results}
\end{table*}

In~\Cref{tab:carla_aggregated_results} we report the detailed object detection performance on all $20$ images under various attacks. 
First of all, the model performs similarly (AP50 around $44\%$) when given three types of benign images: no patch, the patch of a green screen, and the patch of a benign city art painting (\Cref{fig:carla_imperceptible_patches}).
Adding $\ell_\infty$-bounded perturbations generated by our STE-augmented attacks to the city art painting degrades the object detection performance significantly.
The most constrained $\ell_\infty=2$ attack lowers AP50 from $43.29\%$ to $33.20\%$.
In contrast, the unbounded attack without STE is less powerful than our weakest STE-augmented $\ell_\infty=2$ attack,
which only lowers AP50 to $38.73\%$.
Therefore, we skip the experiments of non-STE attacks with $\ell_\infty$ bounds.

Our attack with $\ell_\infty=4$ further lowers AP50 to $17.12\%$.
Subsequently, with $\ell_\infty=8$, AP50 drops to $4.22\%$.
The TIDE error counts suggest that our attacks introduce many hallucinations (TIDE background errors), and cause many object detection misses (TIDE missed errors).
For example, the $\ell_\infty=8$ attack introduces $95$ hallucinations ($15 \rightarrow 110$) and causes $130$ objects to vanish ($306 \rightarrow 436$).
The unbounded STE-augmented attack completely defeats the target model ($0\%$ AP50) because the highly confident adversarial hallucinations overwhelm the detection of other objects that are less confident.

\subsection{Discussion and Future Work}
We have demonstrated that combining \ac{STE} with differentiable rendering produces imperceptible adversarial perturbations to rectangular patches.
One natural extension is to produce imperceptible adversarial camouflage for 3D objects in the physical world.
Since differentiable renderers that support 3D meshes, such as PyTorch3D~\cite{ravi2020accelerating}, are readily available,
users should be able to produce imperceptible adversarial textures using our approach~\Cref{fig:ste_method_patch}.
The other interesting direction is to combine with \ac{EoT}, so that we can make imperceptible adversarial examples in the physical world that are robust to environmental fluctuations.

%% file: 5.conclusion.tex
\section{Conclusion}
In this work, we show that our \ac{STE}-augmented attacks against \ac{DNN}-based visual sensing systems with non-differentiable distortions are effective:
\begin{enumerate*}[1)]
    \item they force zero classification accuracy in the global perturbation threat model;
    \item cause near zero AP50 ($4.22\%$) in object detection in the patch perturbation threat model.
\end{enumerate*}
In contrast to existing physically realizable adversarial examples, our approach produces imperceptible adversarial examples,
making it harder to detect them in the physical world.
This necessitates further investigation into the efficacy of our method, and its implications for real-world systems.
Simultaneously, it will be crucial to evaluate existing defense solutions against such attacks to better understand the landscape of the threat.

%% file: 6.ack.tex
\section*{Acknowledgments} 
This work is partially supported by the Defense Advanced Research Projects Agency (DARPA) under Contract No. HR001119S0026.

%% file: 7.appendix.tex
\noindent\begin{minipage}{\textwidth}
    \centering
    \includegraphics[width=0.75\textwidth]{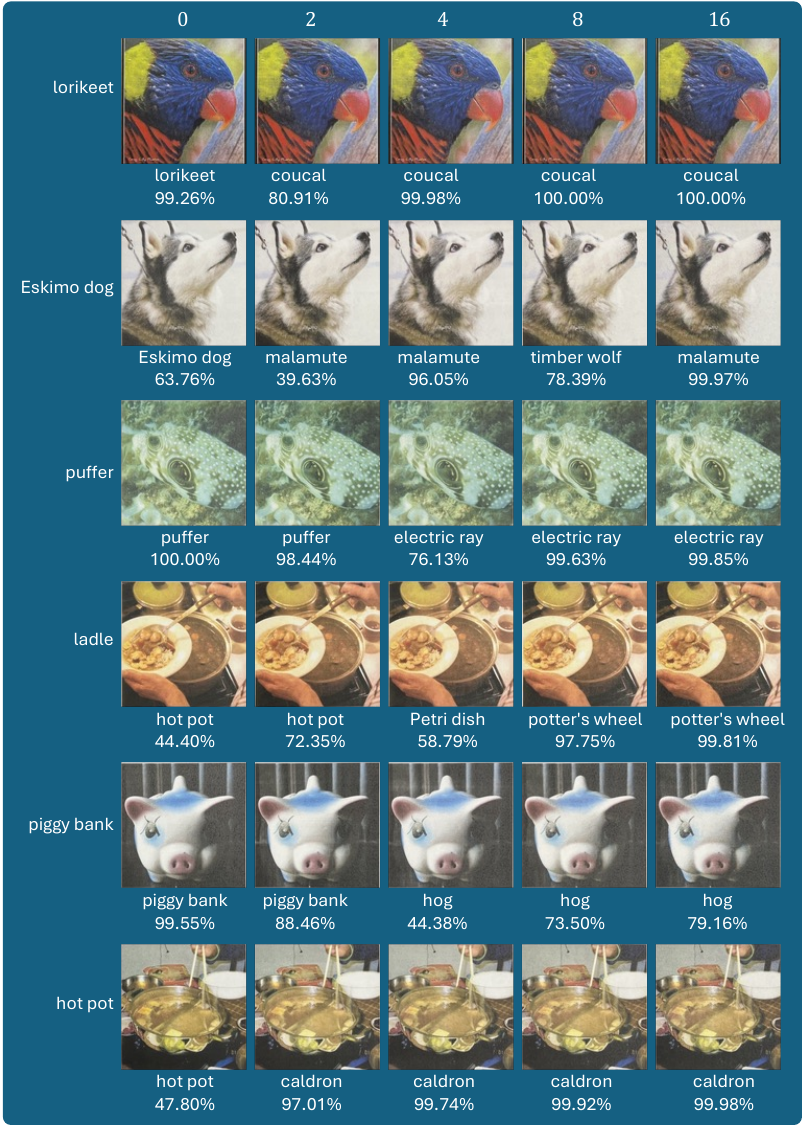}
    \captionof{figure}{Results of the 12-step \ac{PGD} attack combined with \ac{STE} on photos of printouts.
    We annotate the ground truth on the left, and the $\ell_\infty$ bounds at the top. The prediction of the target ResNet50 model is presented below the image. The white edges (most obvious in the ``piggy bank'' row) are the result of imperfect cropping of the photos.}
    \label{fig:pgd_ste_printed_all}
\end{minipage}

\begin{figure*}[tp]
    \centering
    \includegraphics[width=0.75\textwidth]{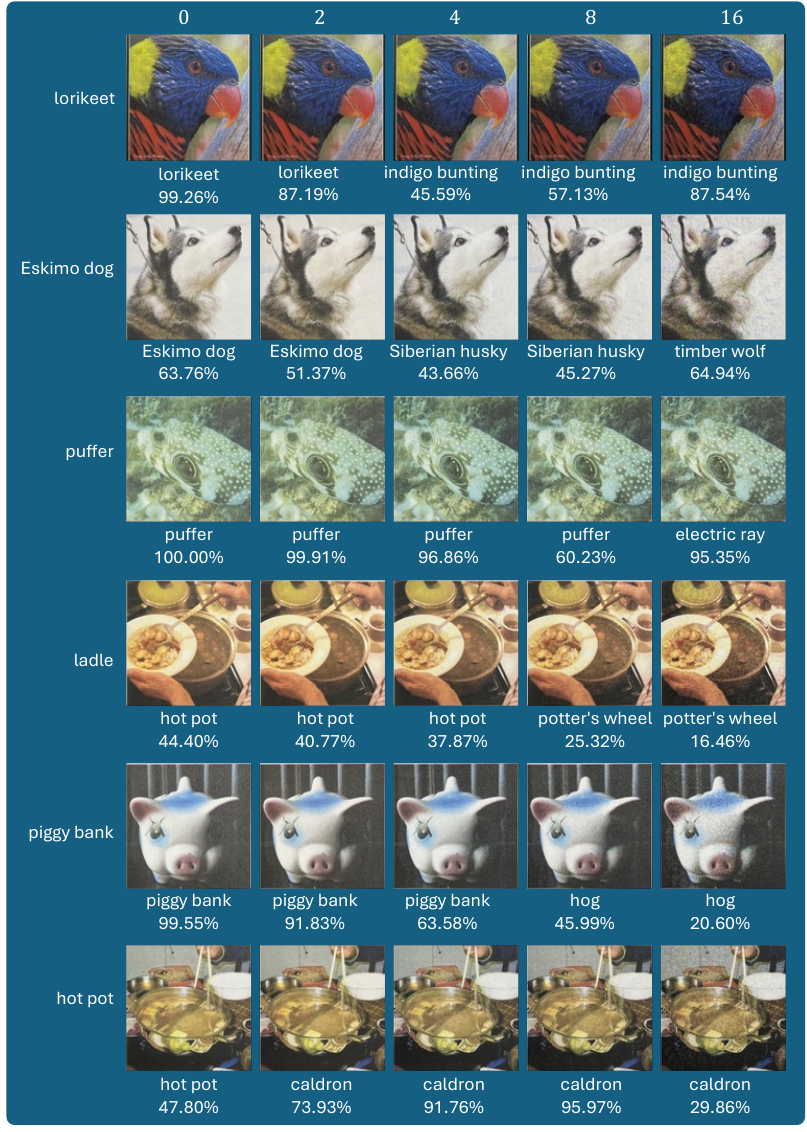}
    \caption{Results of the \ac{FGSM} attack combined with \ac{STE} on photos of printouts.
    We annotate the ground truth on the left, and the $\ell_\infty$ bounds at the top. The prediction of the target ResNet50 model is presented below the image. The white edges (most obvious in the ``piggy bank'' row) are the result of imperfect cropping of the photos.}%
    \label{fig:fgsm_ste_printed_all}
\end{figure*}

\begin{figure*}[tp]
    \centering
    \includegraphics[width=\textwidth]{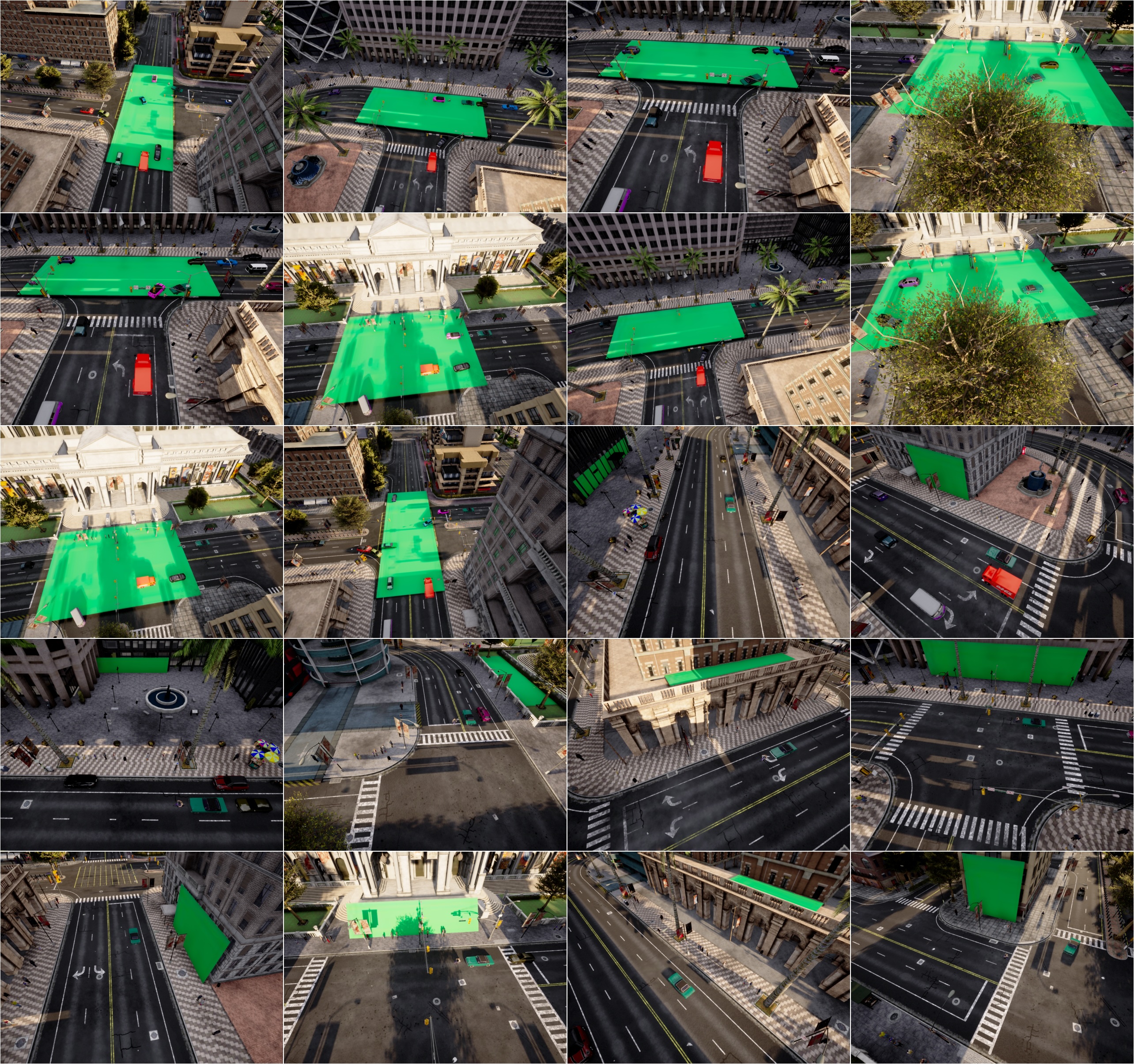}
    \caption{We recreated the scenes of all $20$ images of the DARPA GARD \nth{7} evaluation dataset for object detection in order to render adversarial patches on the green screens in CARLA. We use the bottom right scene to demonstrate imperceptible adversarial patches in the paper.}%
    \label{fig:carla_eval7_dataset}
\end{figure*}

\begin{figure*}[tp]
    \centering
    \includegraphics[width=\textwidth]{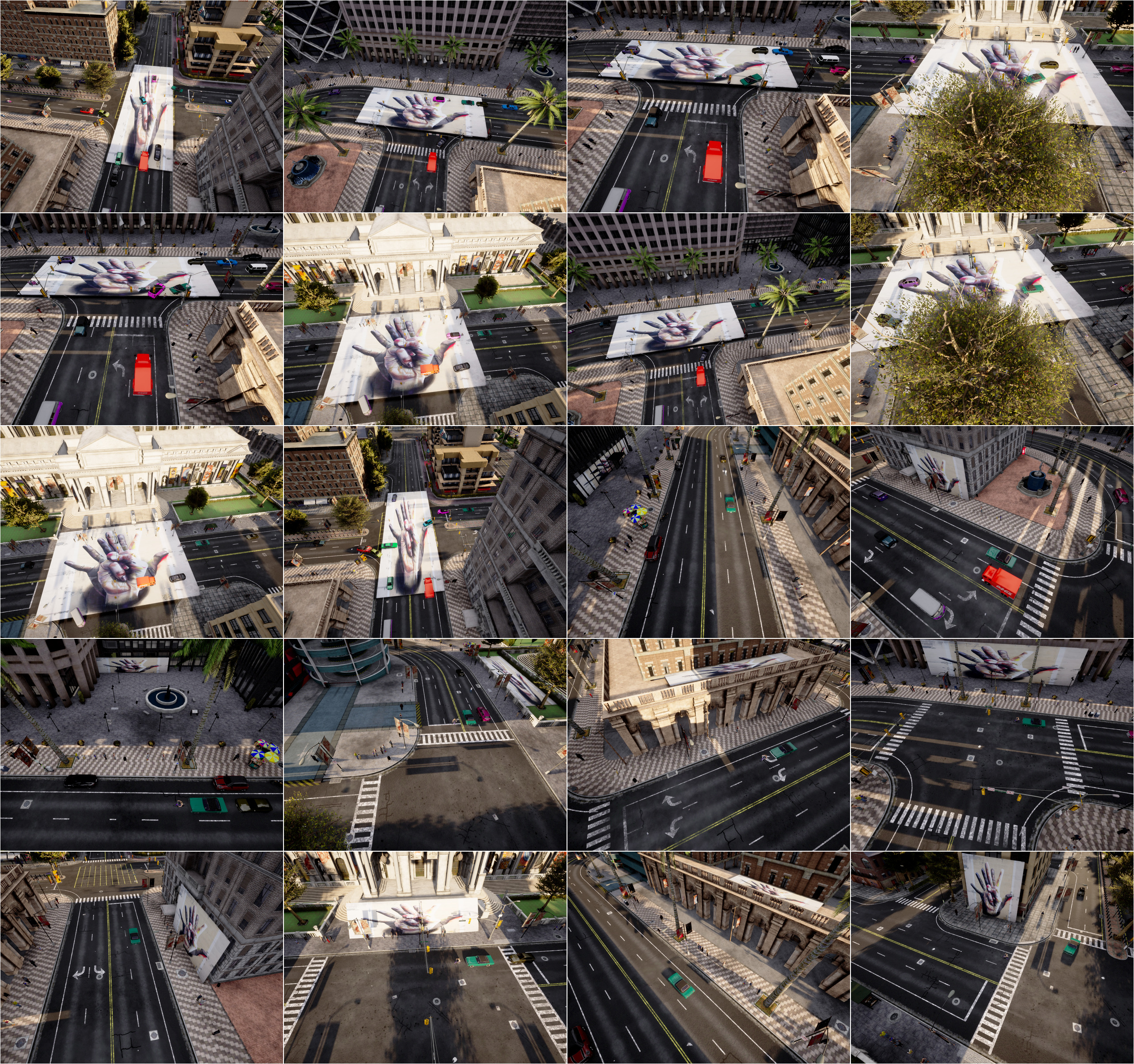}
    \caption{Rendering the benign city art patch in CARLA.}%
    \label{fig:carla_ste_eps0_montage}
\end{figure*}

\begin{figure*}[tp]
    \centering
    \includegraphics[width=\textwidth]{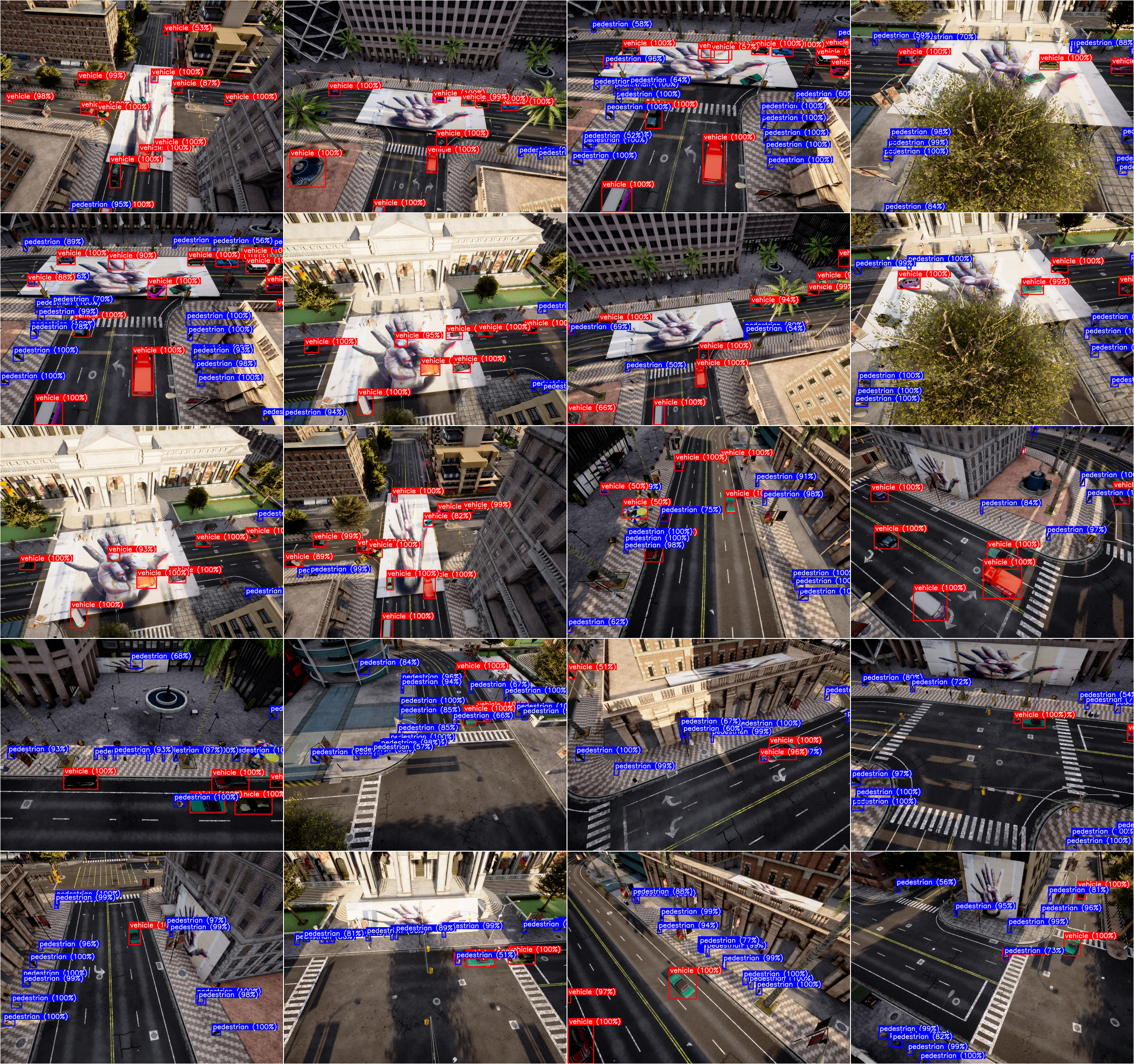}
    \caption{Predictions of the target Faster R-CNN model in CARLA, with the benign city art patch.}%
    \label{fig:carla_ste_eps0_pred_montage}
\end{figure*}

\begin{figure*}[tp]
    \centering
    \includegraphics[width=\textwidth]{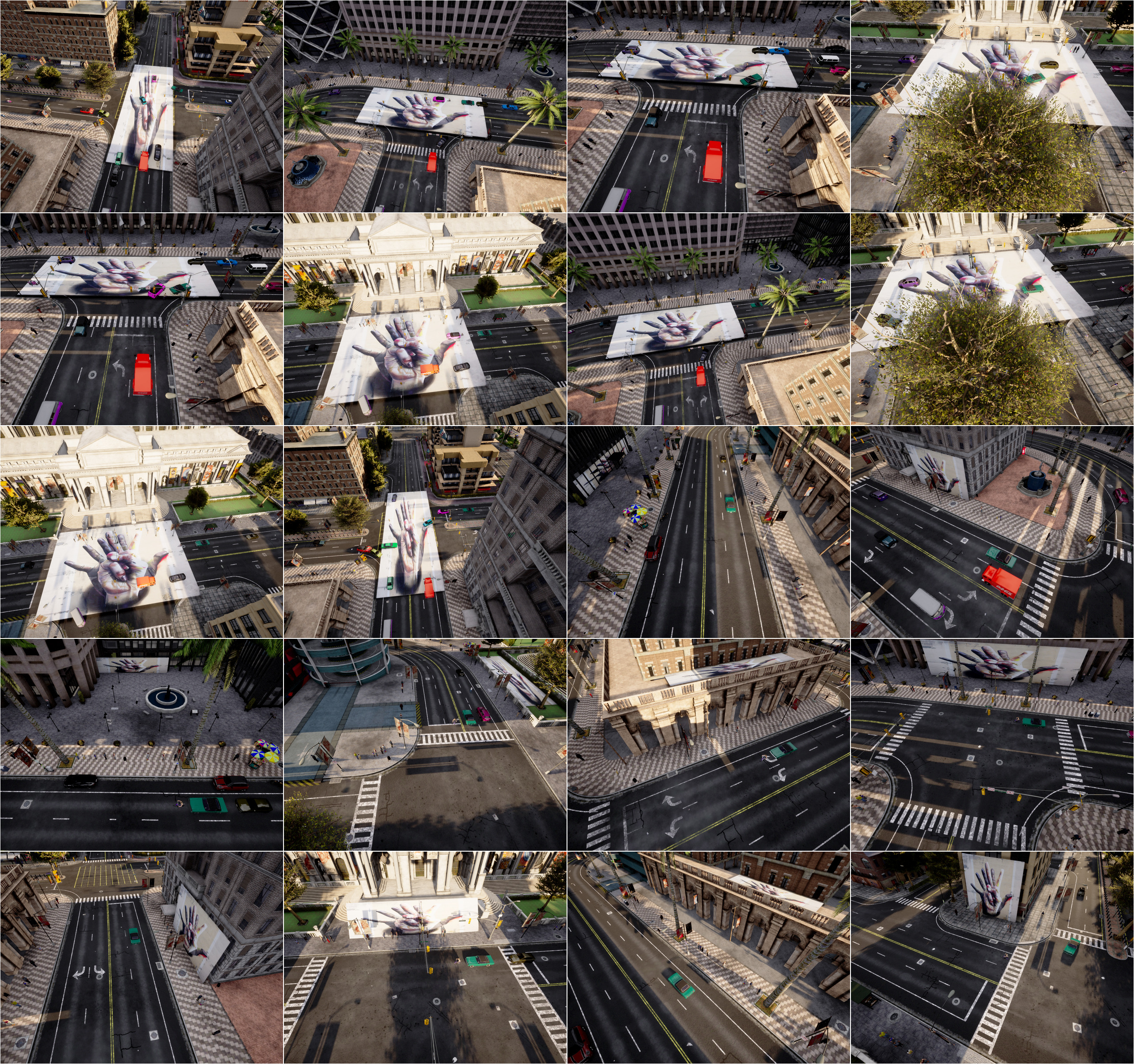}
    \caption{Rendering adversarial patches generated by the \ac{STE}-augmented attack bounded by $\ell_\infty=2$ in CARLA.}%
    \label{fig:carla_ste_eps2_montage}
\end{figure*}

\begin{figure*}[tp]
    \centering
    \includegraphics[width=\textwidth]{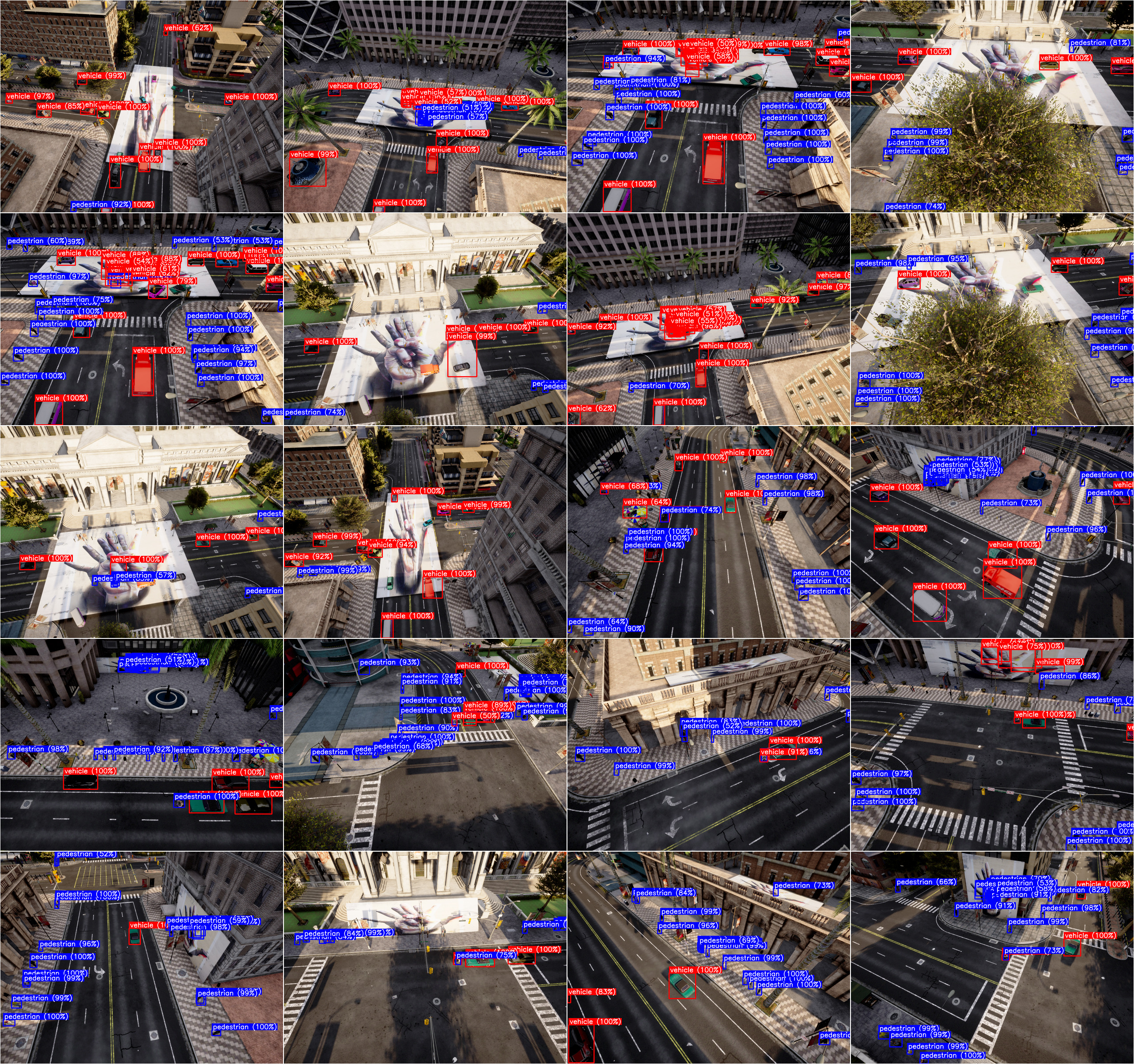}
    \caption{Predictions of the target model in CARLA, with adversarial patches generated by the $\ell_\infty=2$ \ac{STE}-augmented attack.}%
    \label{fig:carla_ste_eps2_pred_montage}
\end{figure*}

\begin{figure*}[tp]
    \centering
    \includegraphics[width=\textwidth]{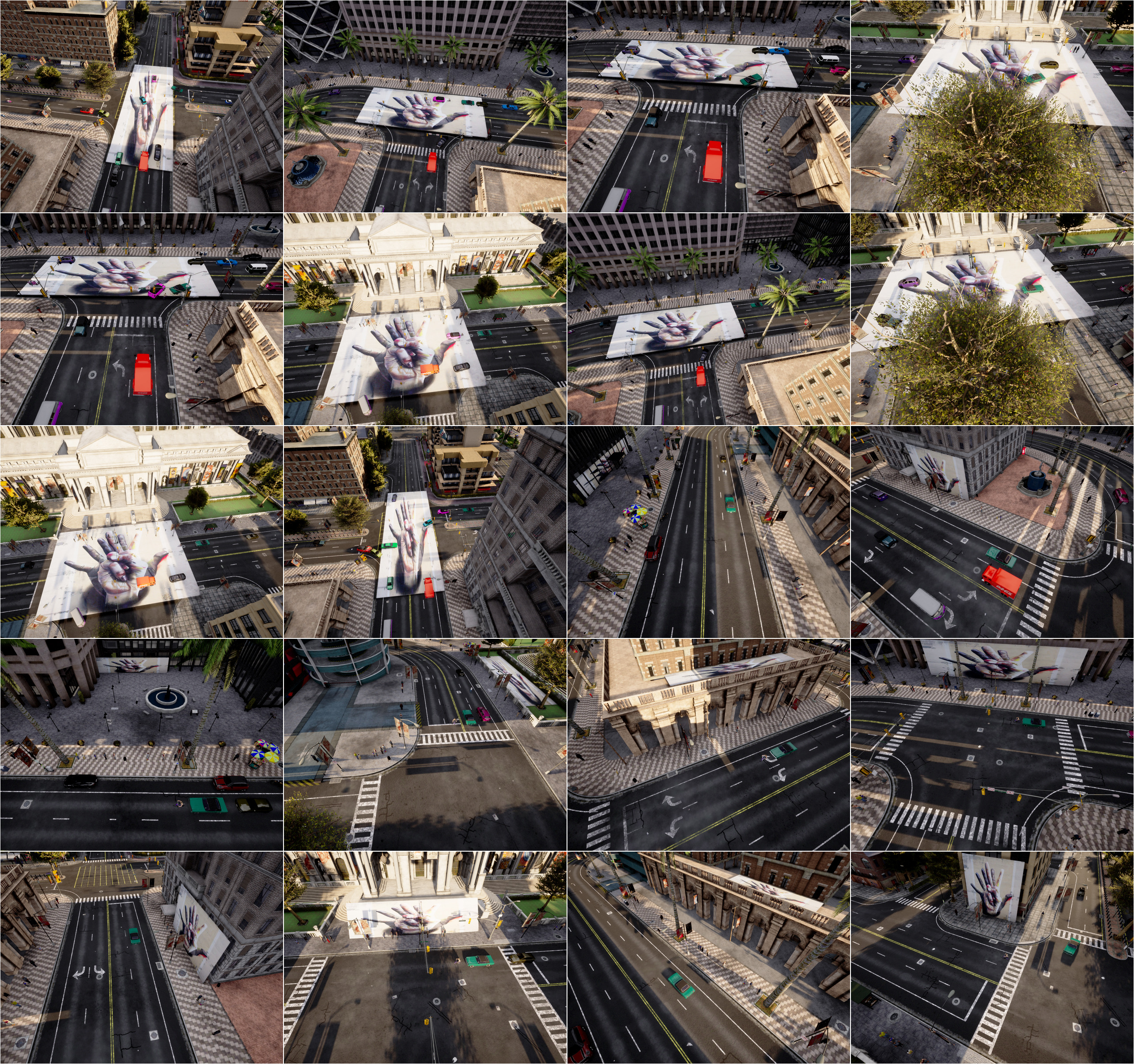}
    \caption{Rendering adversarial patches generated by the \ac{STE}-augmented attack bounded by $\ell_\infty=4$ in CARLA.}%
    \label{fig:carla_ste_eps4_montage}
\end{figure*}

\begin{figure*}[tp]
    \centering
    \includegraphics[width=\textwidth]{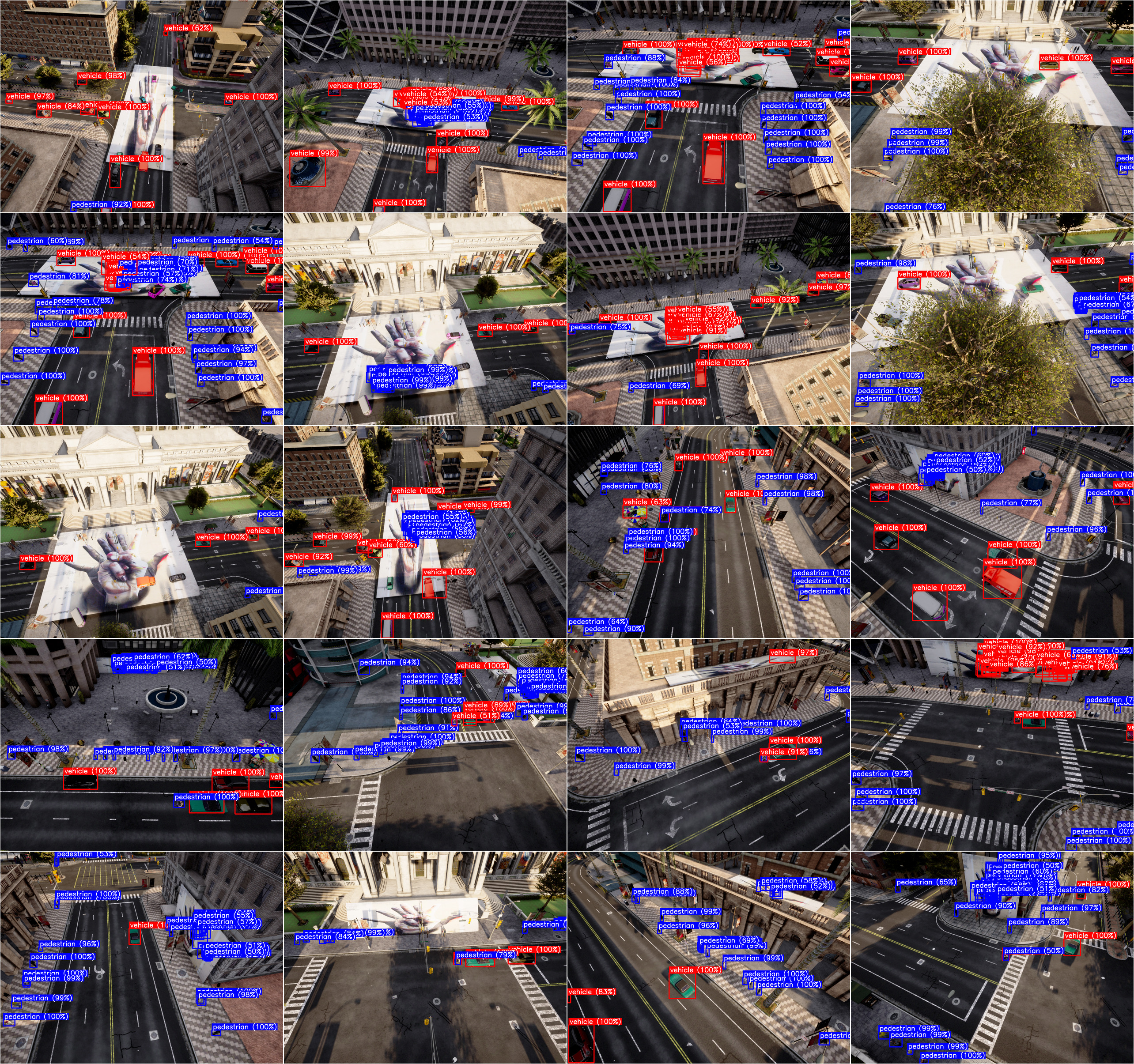}
    \caption{Predictions of the target model in CARLA, with adversarial patches generated by the $\ell_\infty=4$ \ac{STE}-augmented attack.}%
    \label{fig:carla_ste_eps4_pred_montage}
\end{figure*}

\begin{figure*}[tp]
    \centering
    \includegraphics[width=\textwidth]{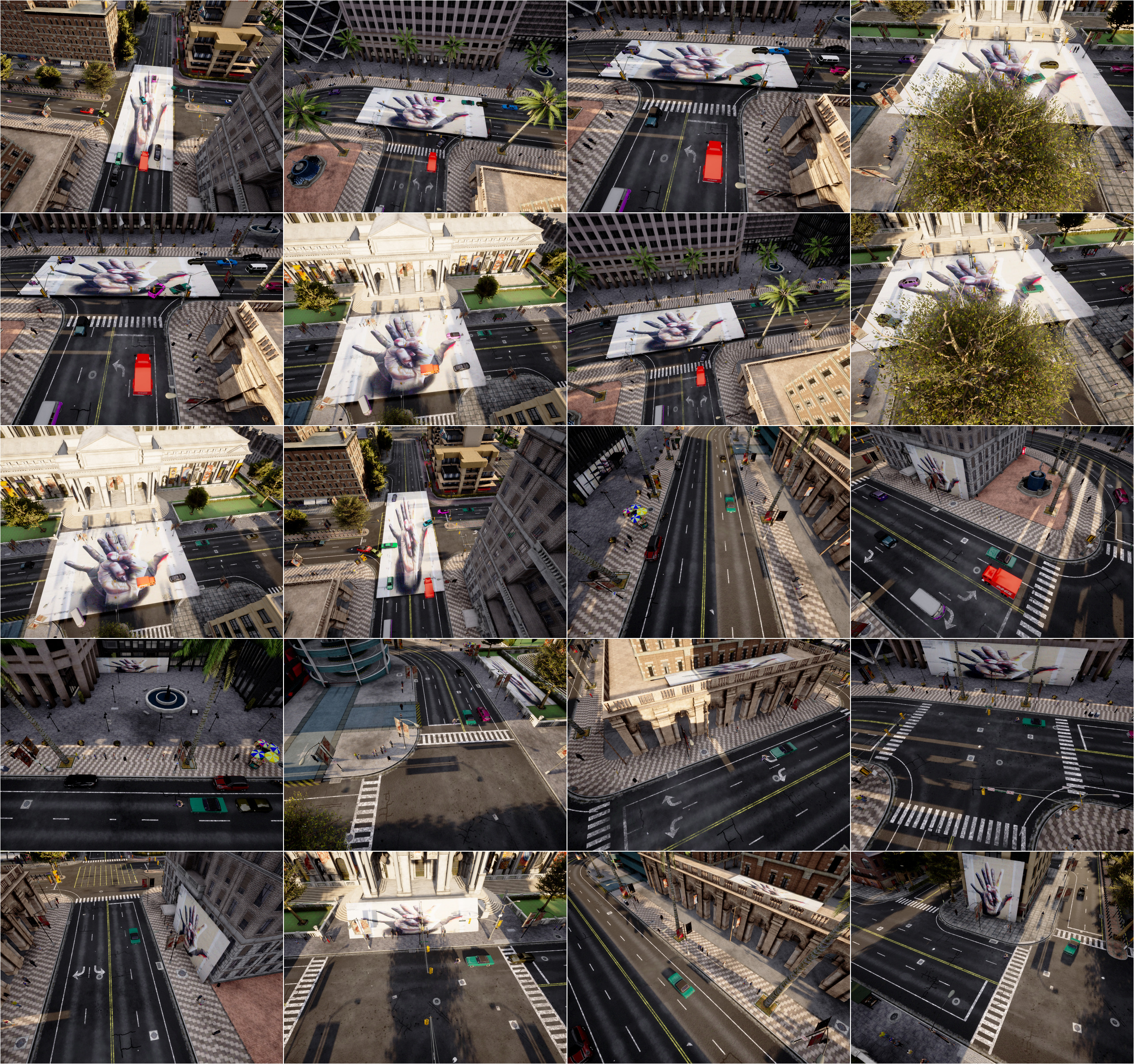}
    \caption{Rendering adversarial patches generated by the \ac{STE}-augmented attack bounded by $\ell_\infty=8$ in CARLA.}%
    \label{fig:carla_ste_eps8_montage}
\end{figure*}

\begin{figure*}[tp]
    \centering
    \includegraphics[width=\textwidth]{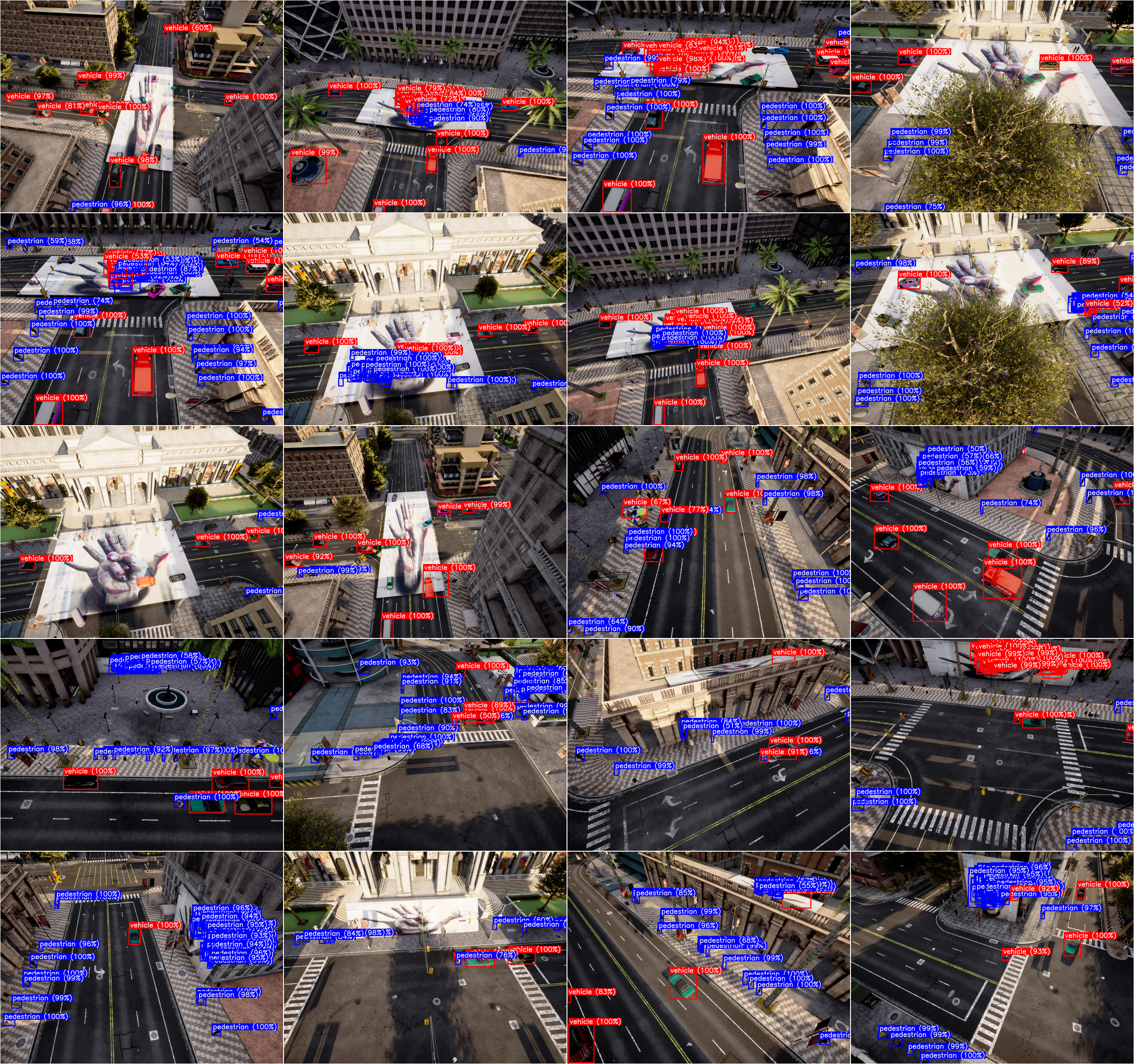}
    \caption{Predictions of the target model in CARLA, with adversarial patches generated by the $\ell_\infty=8$ \ac{STE}-augmented attack.}%
    \label{fig:carla_ste_eps8_pred_montage}
\end{figure*}

\begin{figure*}[tp]
    \centering
    \includegraphics[width=\textwidth]{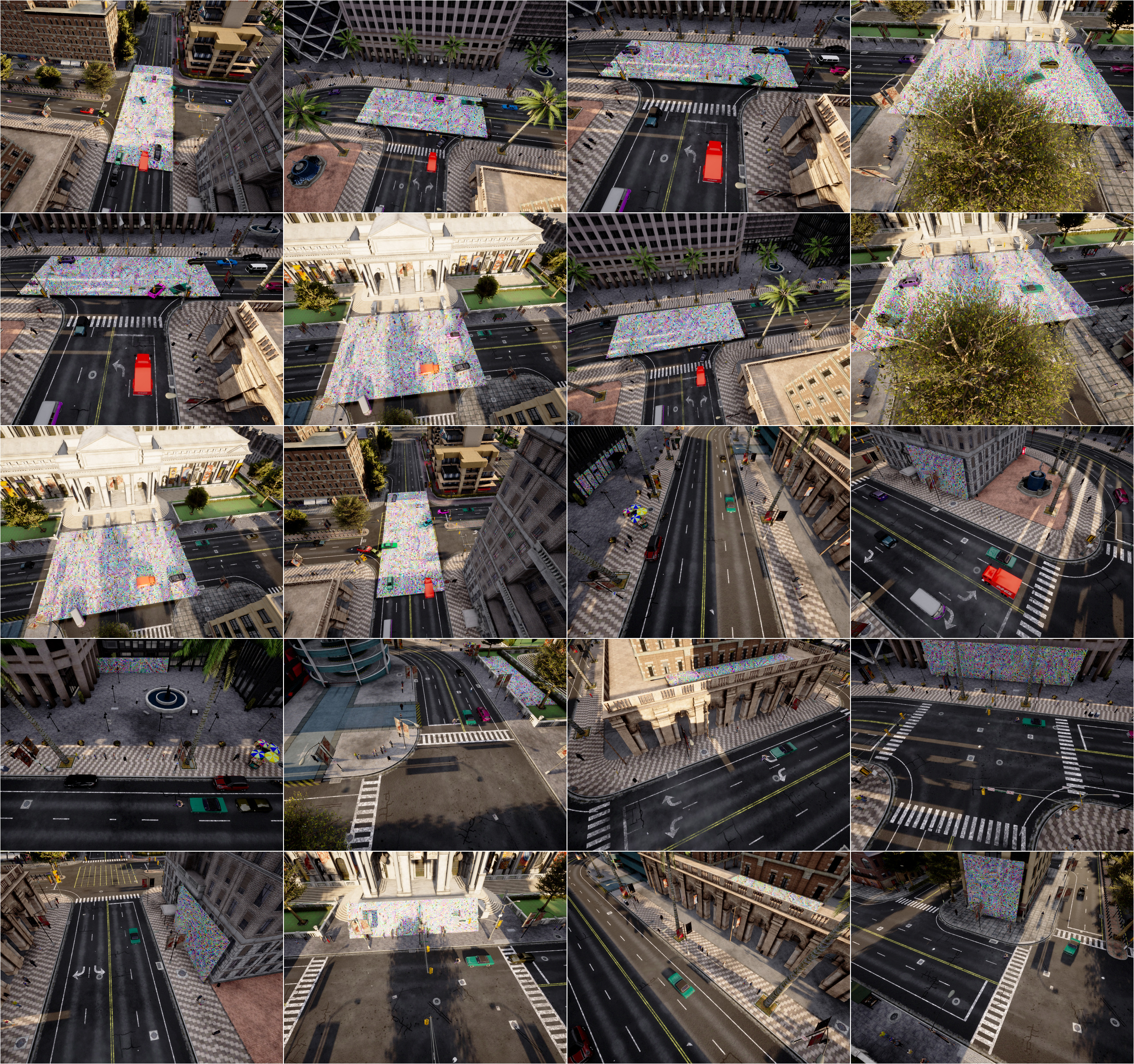}
    \caption{Rendering adversarial patches generated by the unbounded \ac{STE}-augmented attack in CARLA.}%
    \label{fig:carla_ste_eps255_montage}
\end{figure*}

\begin{figure*}[tp]
    \centering
    \includegraphics[width=\textwidth]{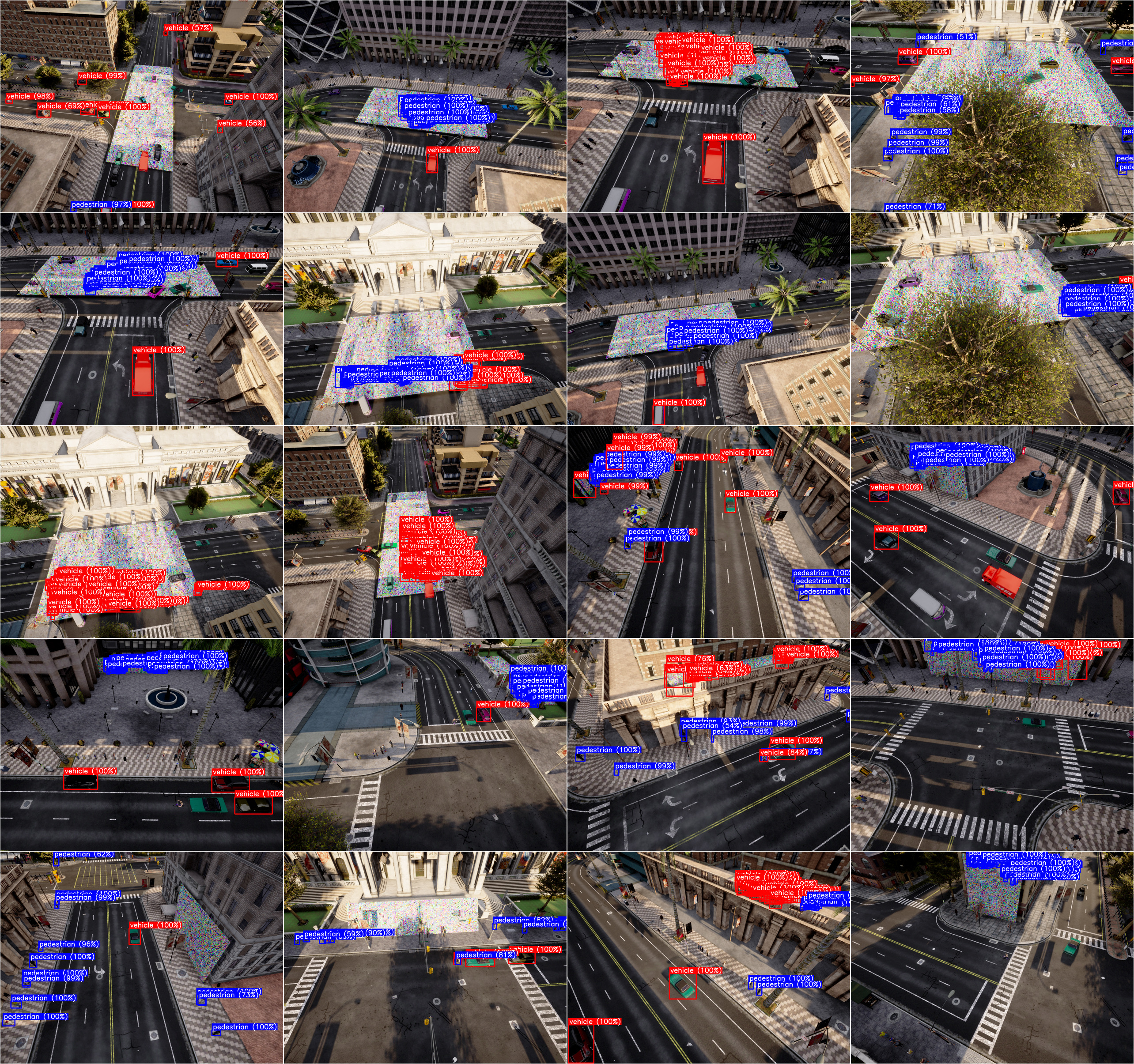}
    \caption{Predictions of the target model in CARLA, with unbounded adversarial patches generated by the \ac{STE}-augmented attack.}%
    \label{fig:carla_ste_eps255_pred_montage}
\end{figure*}